\documentclass[preprint,12pt]{elsarticle}

\usepackage{amssymb}
\usepackage{amsmath}

\usepackage{xcolor}
\usepackage{subcaption}
\usepackage{listings}
\usepackage{epsfig}
\usepackage{graphicx}
\usepackage{color}
\usepackage{tabularx}
\usepackage{colortbl}
\usepackage{amsmath}
\usepackage{pifont}
\usepackage{enumitem}
\usepackage{url}

\journal{IEEE Transactions on Field Robotics}

\makeatletter
\def\ps@pprintTitle{%
 \let\@oddhead\@empty
 \let\@evenhead\@empty
 \renewcommand{\@oddfoot}{\hfil\thepage\hfil} 
 \renewcommand{\@evenfoot}{\hfil\thepage\hfil} 
}
\makeatother

\begin{document}

\begin{frontmatter}



\title{Development of CPS Platform for Autonomous Construction}


\author[1]{Yuichiro Kasahara\corref{cor1}}
\author[1]{Kota Akinari}
\author[1]{Tomoya Kouno}
\author[1]{Noriko Sano}
\author[2]{Taro Abe}
\author[2]{Genki Yamauchi}
\author[2]{Daisuke Endo}
\author[2]{Takeshi Hashimoto}
\author[3]{Keiji Nagatani}
\author[1]{and Ryo Kurazume}

\affiliation[1]{organization={Kyushu University},
            addressline={744 Motooka, Nishi-ku}, 
            city={Fukuoka-shi},
            postcode={819-0395}, 
            state={Fukuoka},
            country={Japan}}

\affiliation[2]{organization={Public Works Research Institute (PWRI)},
            addressline={1-6 Minamihara}, 
            city={Tsukuba-shi},
            postcode={305-8516}, 
            state={Ibaraki},
            country={Japan}}

\affiliation[3]{organization={University of Tsukuba},
            addressline={1-1-1 Tennodai}, 
            city={Tsukuba-shi},
            postcode={305-8573}, 
            state={Ibaraki},
            country={Japan}}
            
\cortext[cor1]{Corresponding author. Email: kasahara.yuichiro.res@gmail.com, Postal addree: Room 928, Building West-2, 744 Motooka, Nishi-ku, Fukuoka 819-0395, Japan, Phone number: +8180-5153-7875}

\begin{abstract}
In recent years, labor shortages due to the declining birthrate and aging population have become significant challenges at construction sites in developed countries, including Japan. To address these challenges, we are developing an open platform called ROS2-TMS for Construction, a Cyber-Physical System (CPS) for construction sites, to achieve both efficiency and safety in earthwork operations. In ROS2-TMS for Construction, the system comprehensively collects and stores environmental information from sensors placed throughout the construction site. Based on these data, a real-time virtual construction site is created in cyberspace. Then, based on the state of construction machinery and environmental conditions in cyberspace, the optimal next actions for actual construction machinery are determined, and the construction machinery is operated accordingly. In this project, we decided to use the Open Platform for Earthwork with Robotics and Autonomy (OPERA), developed by the Public Works Research Institute (PWRI) in Japan, to control construction machinery from ROS2-TMS for Construction with an originally extended behavior tree. In this study, we present an overview of OPERA, focusing on the newly developed navigation package for operating the crawler dump, as well as the overall structure of ROS2-TMS for Construction as a Cyber-Physical System (CPS). Additionally, we conducted experiments using a crawler dump and a backhoe to verify the aforementioned functionalities.
\end{abstract}


\begin{keyword}


Cyber Physical System (CPS), OPERA, Autonomous Construction, Behavior Tree
\end{keyword}

\end{frontmatter}




\section{Introduction}
\begin{figure}[htb]
\centering
\includegraphics[width=75mm]{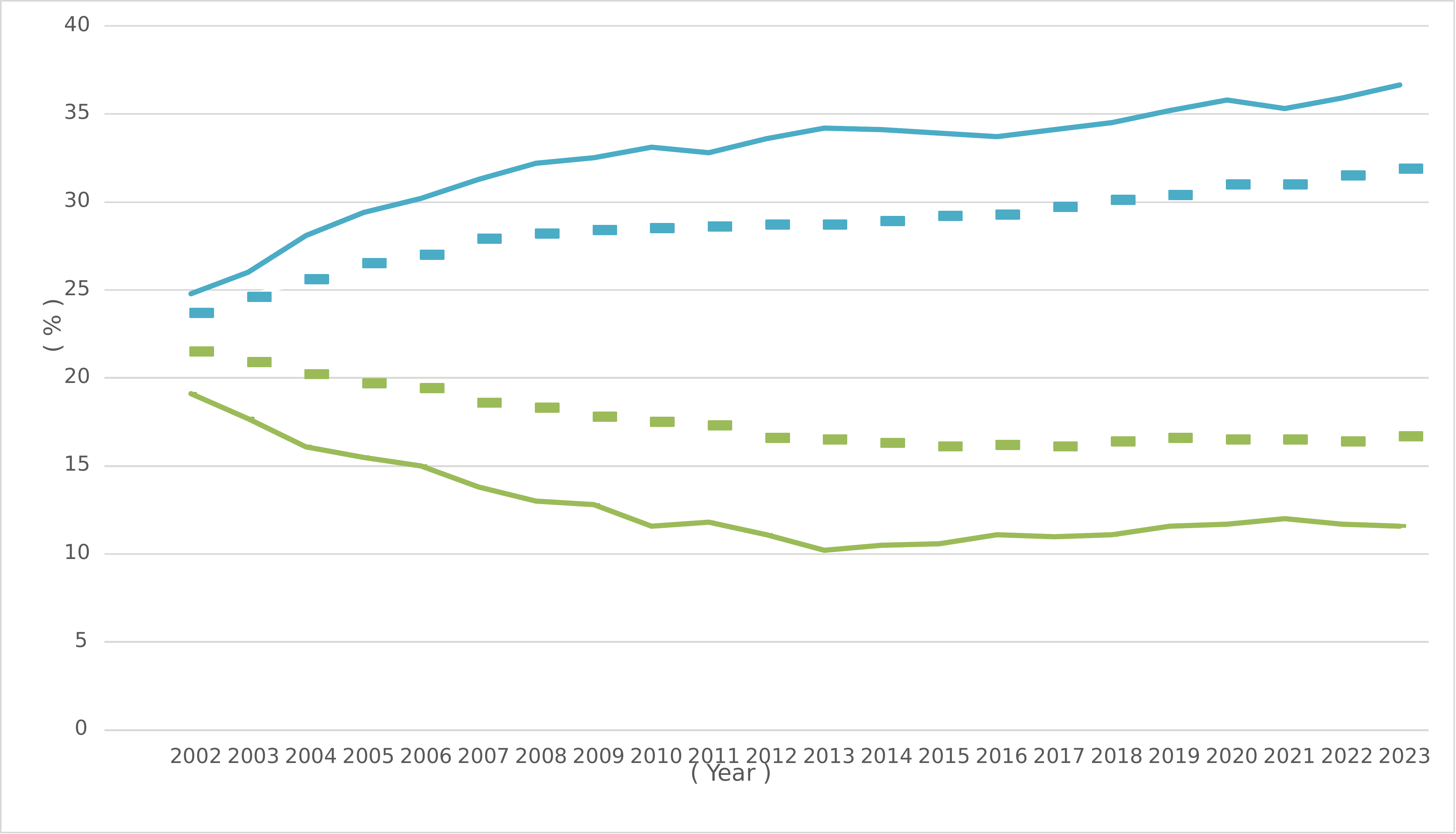}
\caption{The ration of employees in Japan based on ~\cite{estat_1}}
\label{fig:ratio_of_employees}
\end{figure}

\begin{figure}[htb]
\centering
\includegraphics[width=75mm]{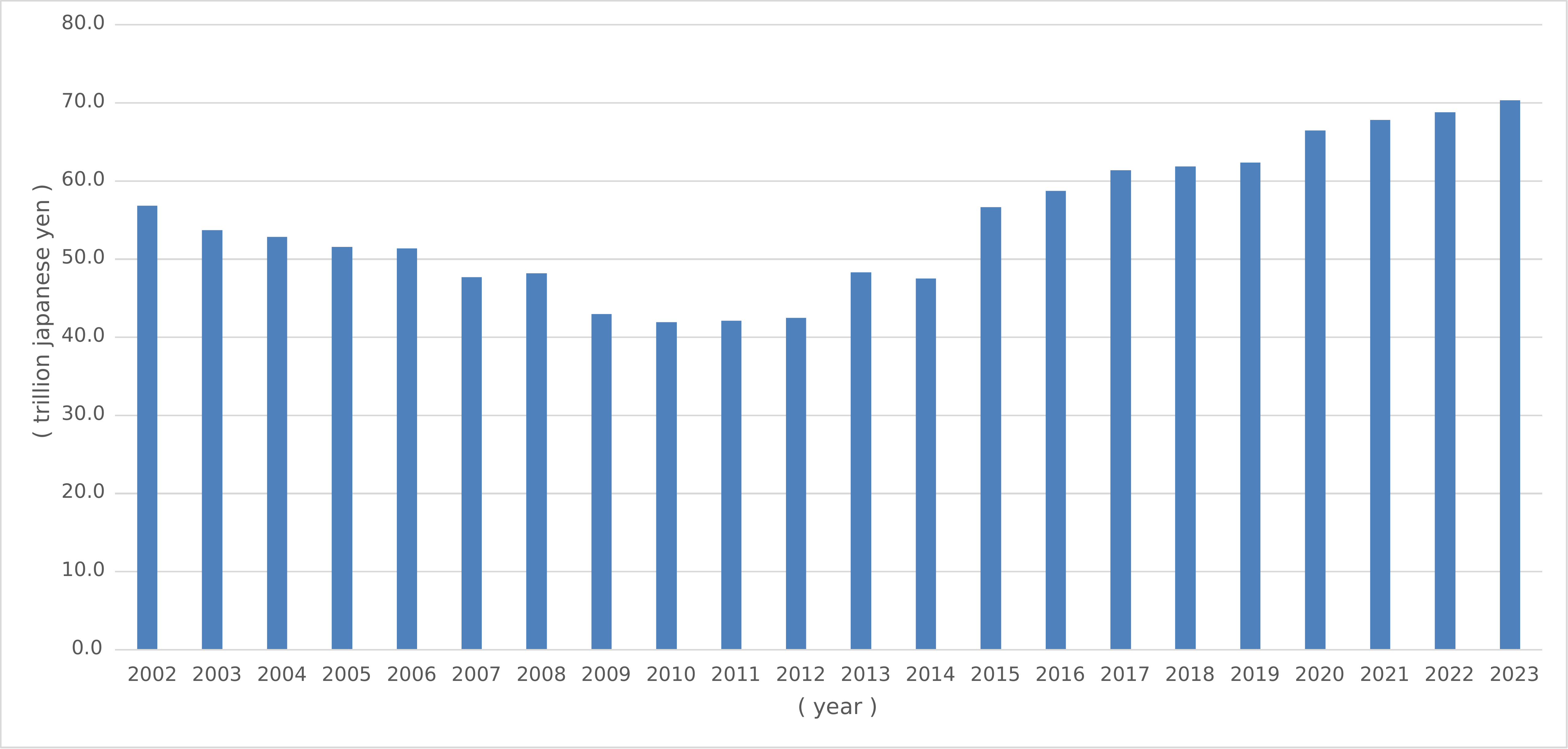}
\caption{The transition of construction investment amount in Japan based on ~\cite{estat_2}}
\label{fig:construction_investment}
\end{figure}

In the near future, Japan is expected to face a serious decline in skilled workers and new recruits at construction sites, as shown in Fig.~\ref{fig:ratio_of_employees}. On the other hand, as shown by the trends in construction investment in Fig.~\ref{fig:construction_investment}, the demand for construction has slightly increased. As a result, the realization of autonomous construction to improve productivity has become an urgent issue. 

Since the launch of i-Construction~\cite{i-Construction} in 2016, efforts such as autonomous construction have been actively promoted in Japan. Among the various staged goals of i-Construction, the final objective of achieving fully autonomous construction has been pursued by many companies, as exemplified by initiatives like A4CSEL~\cite{A4CSEL}, Autonomous Haulage System (AHS)~\cite{AHS_1}~\cite{AHS_2} and others ~\cite{TAISEI}~\cite{CAT}~\cite{kobelco}. However, these systems have been developed independently by each construction company and machinery manufacturer under non-disclosure agreements (NDAs). In such a situation, information sharing both between individual R\&D groups and with external parties becomes weak, leading to issues such as the duplication of technology development and reduced versatility of the technologies. 

To tackle these issues, the Public Works Research Institute (PWRI) proposed common control messages~\cite{common_control_messages}. This is an open standard proposal for CAN signals used in the autonomous operation of construction machinery. To support the adoption of this proposal, its practical implementation and related applications were publicly released as the OPERA~\cite{opera}. However, OPERA lacks task scheduling functionality, and in order to achieve autonomous construction, it was necessary to build a system with this capability on OPERA. Therefore, we developed ROS2-TMS for Construction~\cite{git_ros2_tms_for_construction}, a CPS platform for construction sites, which is the first system to control actual construction machinery using OPERA. 

\begin{figure}[htb]
\centering
\includegraphics[width=75mm]{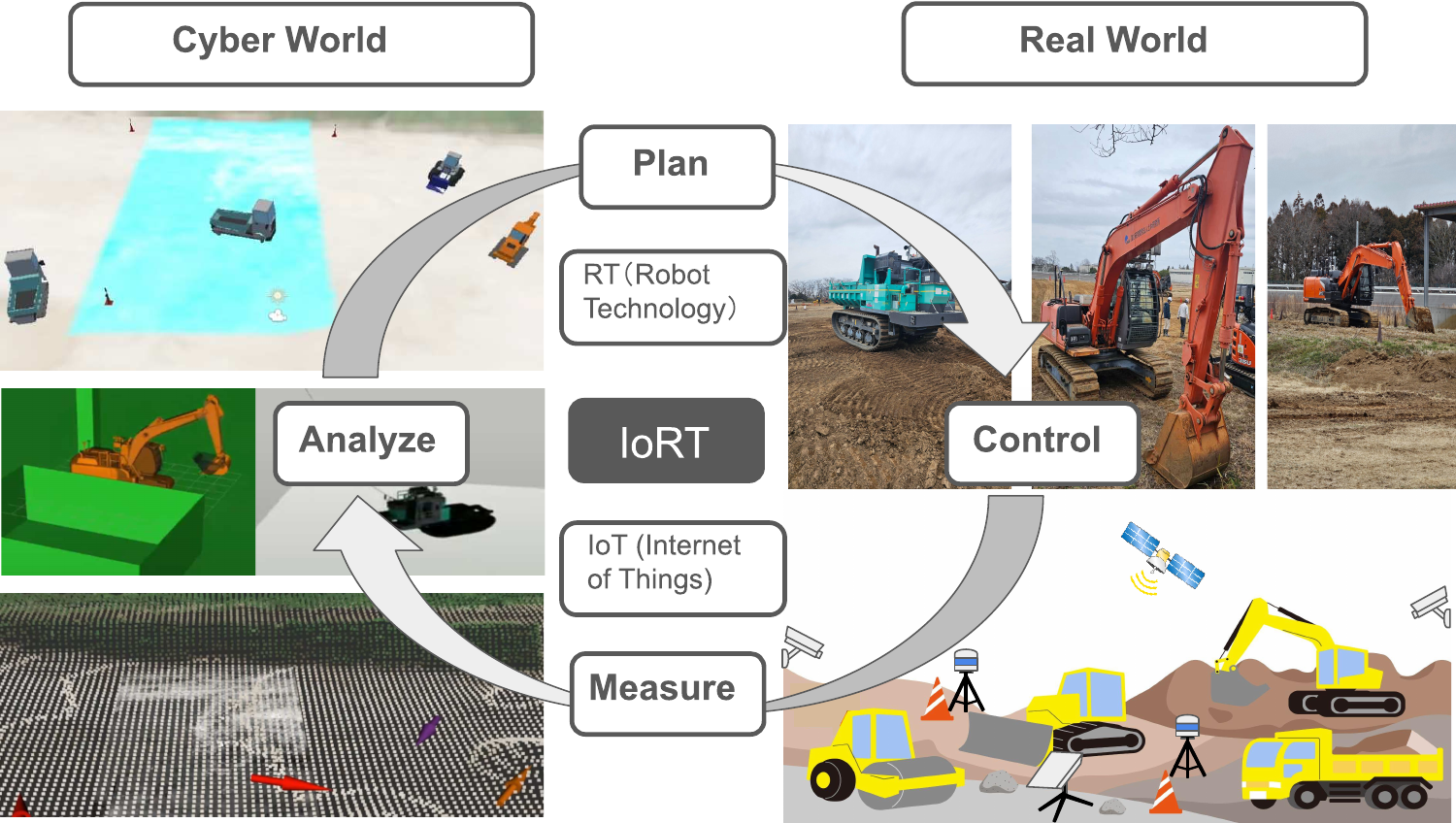}
\caption{The concept of ROS2-TMS for Construction}
\label{fig:ros_tms_for_construction}
\end{figure}

ROS-TMS~\cite{ros_tms}\cite{ros2_tms}, the predecessor of ROS2-TMS for Construction~\cite{git_ros2_tms_for_construction}, is a system based on the concept of an Informationally Structured Environment~\cite{informationally_structured_environment}, where sensors are placed throughout the environment in which robots operate. These sensors collect and analyze data (environmental information) to build a real-time virtual environment in cyberspace, which is then used to determine the optimal actions the robots should take next. As shown in Fig.~\ref{fig:ros_tms_for_construction}, ROS2-TMS for Construction is an adaptation of this system, originally designed for indoor environments, to suit construction sites. The major changes in ROS2-TMS for Construction include the use of Behavior Tree (BT) as the task scheduler and the application of OPERA for controlling the actual construction machinery. In ROS-TMS, the Finite State Machine (FSM) has traditionally been used as a task scheduler. However, with the transition to ROS-TMS for Construction, we have switched to a Behavior Tree (BT). This change was made because tasks created using BT are more clearly defined in terms of starting points, and the tasks themselves offer greater reusability compared to FSM. Additionally, the trend in the robotics industry has seen an accelerated shift from FSM to BT\cite{behaviortree}. On the other hand, BT is less flexible compared to FSM, requiring the necessary operations to be explicitly incorporated into the tasks, which can lead to large and redundant task structures. Furthermore, in tasks involving multiple construction machinery that require synchronization, the conventional Behavior Tree essentially requires combining all machinery into a single task. If irregularities occur while operating multiple machines, it becomes necessary to stop the entire task, including the parts that are unaffected by the issue. To address these shortcomings of previous task management mechanism\cite{ros2_tms_for_construction}, we have developed a global blackboard, the information sharing mechanism that enables synchronization between multiple tasks, aiming to enhance the flexibility of BT. Furthermore, we revised the task design principles towards achieving open design\cite{nagatani_1} in construction sites. In addition to these changes, we also implemented OperaSimVR, a feature that builds a real-time virtual construction site in cyberspace based on environmental information collected from sensors placed throughout the construction site. In this study, we introduce the overall structure of OPERA, focusing on the  newly implemented navigation functionality of crawler dump IC120, as well as the Cyber-Physical System (CPS) architecture of ROS2-TMS for Construction. Additionally, we present the results of validation experiments of aforementioned functionalities conducted using the crawler dump IC120 and the backhoe ZX200.\\

The remainder of this study is organized as follows: 
\begin{itemize}
    \item OPERA: Introduces the scope and functionality of OPERA icluding the newly implemented navigation functionality of the IC120 crawler dump.
    \item ROS2-TMS for Construction: Describes the overall structure of ROS2-TMS for Construction, along with its two major features: the task management mechanism and OperaSimVR. 
    \item Experiments: Describes the results of construction experiments conducted using ROS2-TMS for Construction and OPERA. 
    \item Conclusion: Summarizes the functionalities introduced in this paper and provides an evaluation and outlook for the future.
\end{itemize}

\section{OPERA}

\begin{figure}[htb]
\centering
\includegraphics[width=75mm]{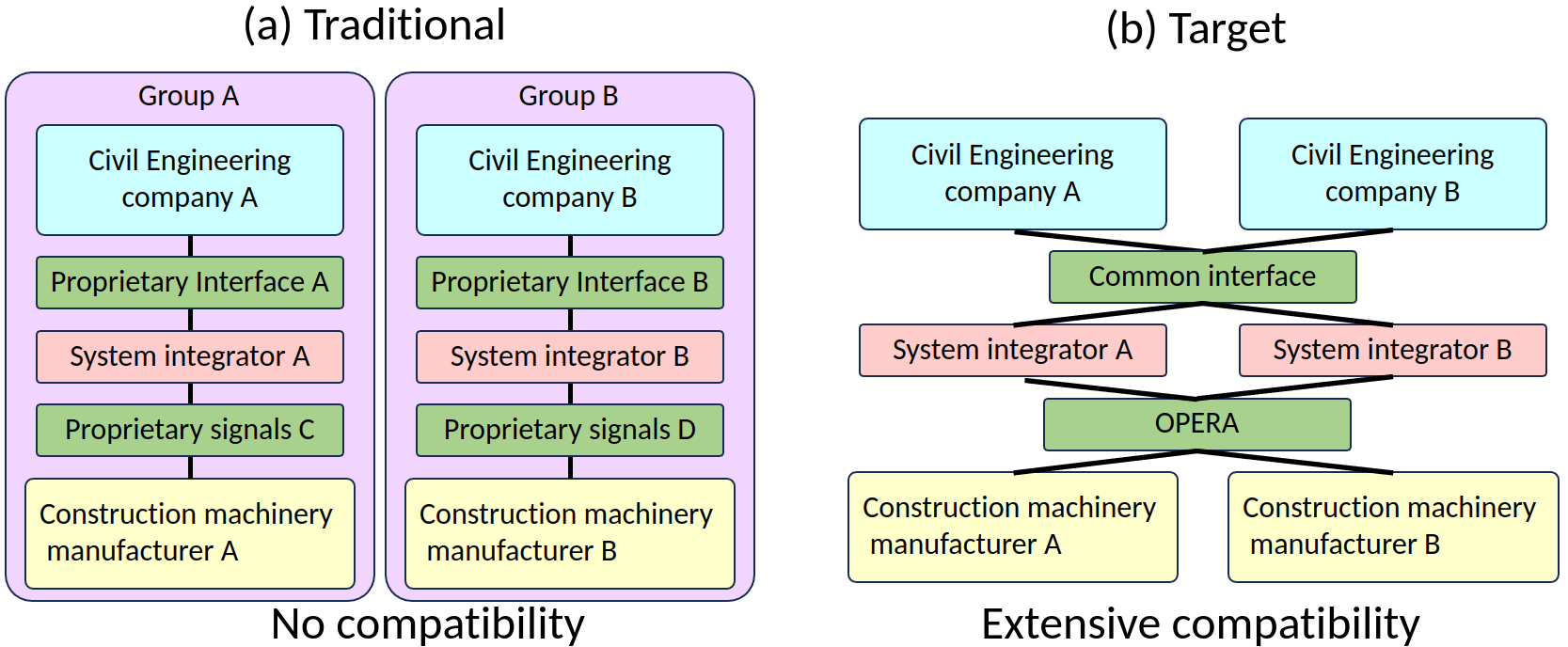}
\caption{The target of OPERA}
\label{fig:OPERA_target}
\end{figure}

In this study, ROS2-TMS for Construction controls actual construction machinery using OPERA, which has been developed by PWRI. As shown in Fig.~\ref{fig:OPERA_target}(a), control signals used for the automation of construction machinery have so far been developed by individual R\&D groups. However, this has led to the following issues.

\begin{itemize}
    \item Overlapping research and development investments by different research groups.
    \item The internal signals of construction machinery are confidential, requiring contracts and NDAs with manufacturers for access. This creates a significant barrier for new entrants, such as universities and Small and Medium Enterprises (SMEs).
    \item Even among similar types of construction machinery, internal signals differ between manufacturers, making it difficult to use the same control system for autonomous operation, which lowers the reusability of developed technologies. 
\end{itemize}

To solve these issues, PWRI proposed a public standard called "common control messages," which are CAN-format signals designed for the autonomous control of construction machinery. By using common control messages, it is possible to unify the signals used for autonomous control of construction machinery, as shown Fig.~\ref{fig:OPERA_target}(b). By using common control messages, the following benefits can be realized. 

\begin{itemize}
    \item Reduction of competitive areas and overlapping development investments in the autonomous control of construction machinery.
    \item Reduction of barriers to entry for new participants in the development of autonomation of construction machinery through the public release of signals.
    \item Even if construction machinery is manufactured by different companies, the use of the same control signals for the same type construction machinery will enable autonomous operation, thereby improving the reusability of the developed autonomous control technologies. 
\end{itemize}

\begin{figure}[htb]
\centering
\includegraphics[width=75mm]{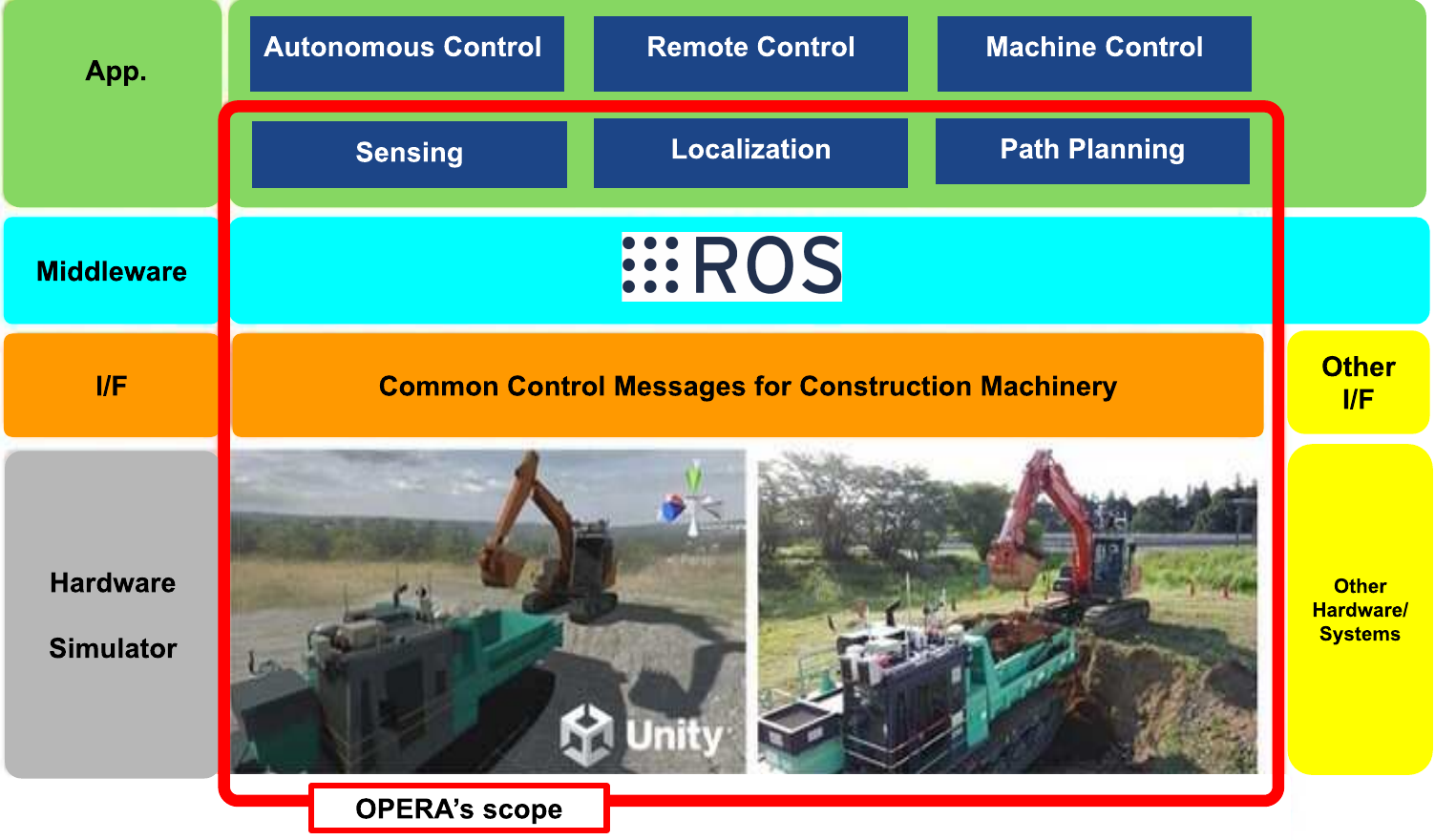}
\caption{OPERA}
\label{fig:opera}
\end{figure}

\begin{figure}[htb]
\centering
\includegraphics[width=75mm]{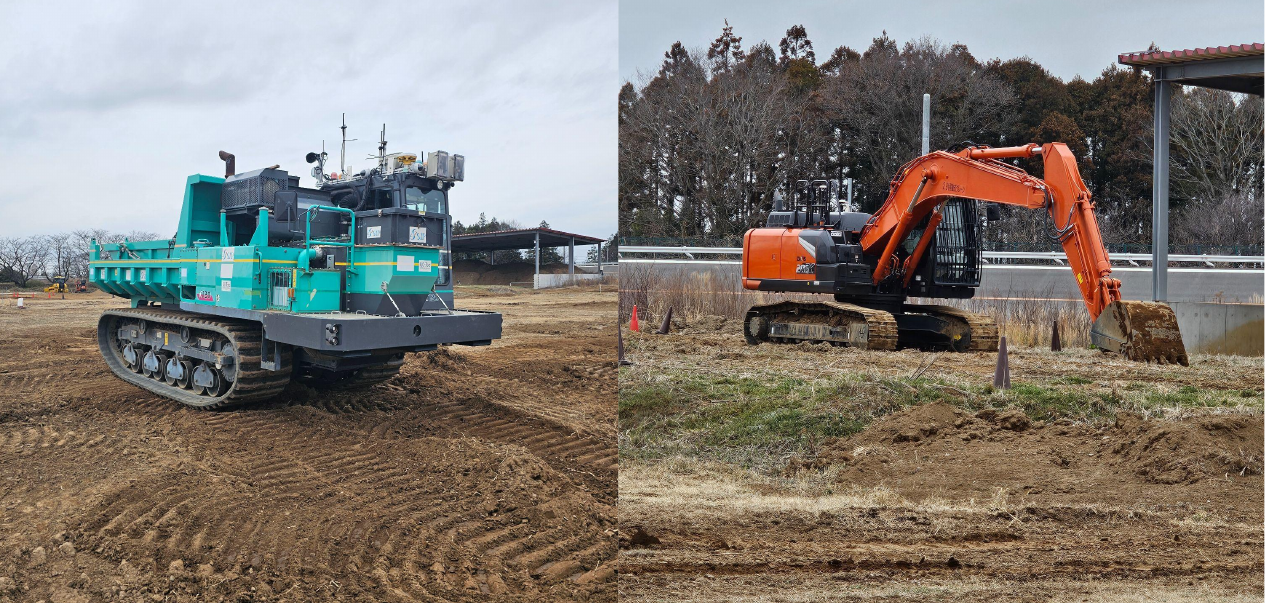}
\caption{OPERA-compatible crawler dump IC120 and backhoe ZX200}
\label{fig:opera_compatible_machinery}
\end{figure}

However, common control messages are merely a signal proposal for the autonomous operation of construction machinery. To enable their practical use, it is necessary to implement the common control messages on actual construction machinery and develop the applications required for autonomous construction. In response, PWRI developed an open platform called OPERA, aimed at the autonomous control of construction machinery using common control messages. As shown in Fig.~\ref{fig:opera}, OPERA consists of actual construction machinery (Common control messages compatible construction machinery and the experimental field), a simulator\cite{opera_simulator} (virtual construction machinery and a virtual experimental field)\cite{operasim_physx}\cite{operasim_agx}, middleware (ROS~\cite{ros}/ROS2~\cite{ros2}), and applications such as navigation and manipulation tools. The actual construction machinery compatible with the common control messages, the crawler dump IC120 and the backhoe ZX200(Henceforth, OPERA-compatible crawler dump IC120 (kato works co. ltd) \cite{ic120} and backhoe ZX200 (Hitachi Construction Machinery Japan Co., Ltd.) \cite{zx200} are simply referred to as IC120 and ZX200), are shown in Fig.~\ref{fig:opera_compatible_machinery}.

In this way, OPERA provides a variety of functionalities, including simulators and actual machinery, to enable users to more easily engage in technology development aimed at the automation of construction machinery. Additionally, OPERA offers the rental of construction machinery and fields to the public, promoting participation in this field by research institutions such as SMEs, universities, and national research institutes that previously faced challenges in entering the field. From here on, we will introduce the navigation module for the IC120 using Navigation2~\cite{nav2} and the manipulation module for the ZX200 using MoveIt!2~\cite{moveit!2}, which were implemented as part of OPERA.

\subsection{Navigation module for crawler dump IC120}
\begin{figure}[htb]
\centering
\includegraphics[width=75mm]{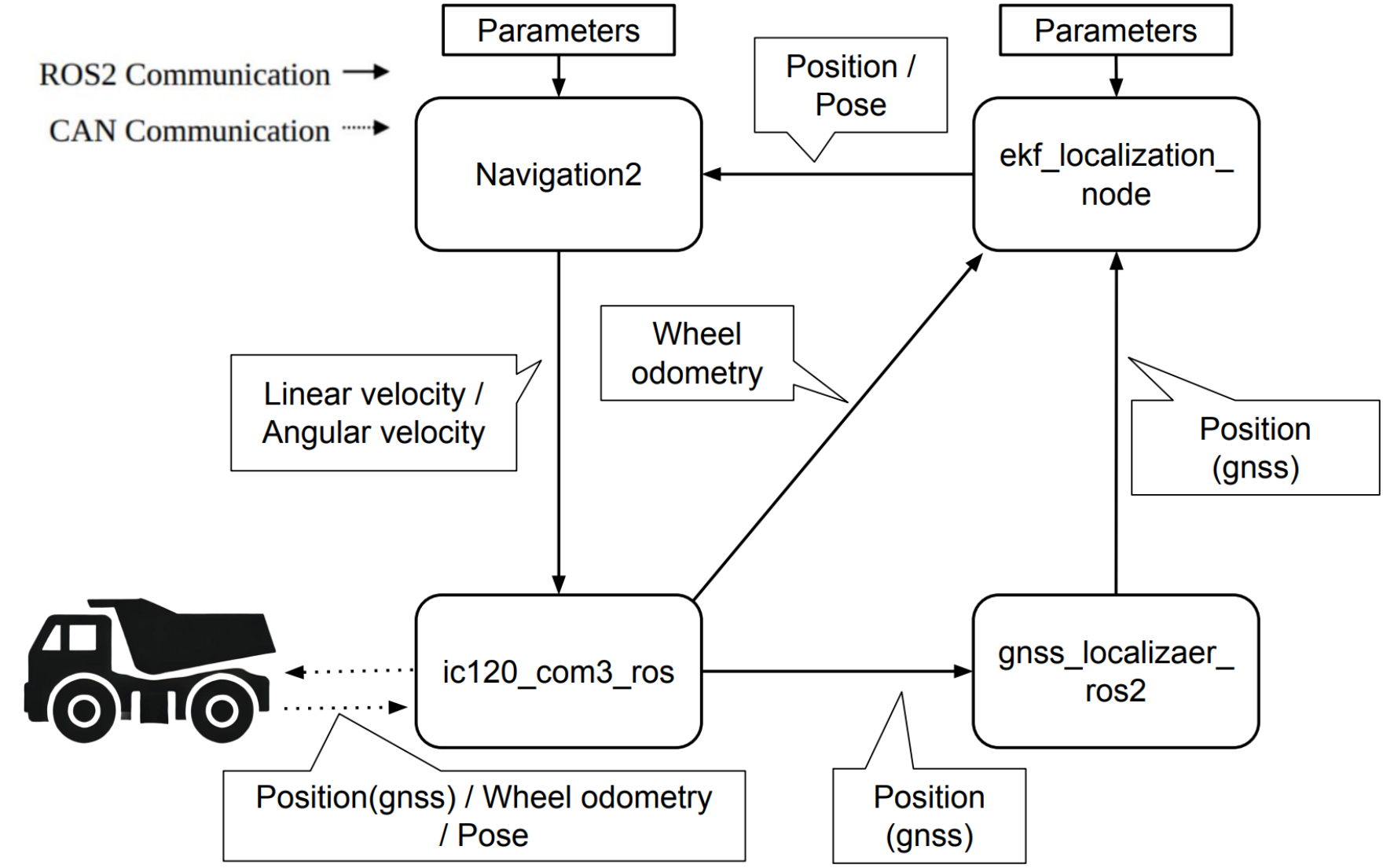}
\caption{The architecture of ic120\_ros2}
\label{fig:ic120ros2_architecture}
\end{figure}

\sloppy
For the navigation of the IC120, we developed a set of ROS2 packages called ic120\_ros2~\cite{ic120_ros2}. In ic120\_ros2, self-position estimation is performed using an extended Kalman filter based on sensor data from the wheel encoders and GNSS installed in the IC120. Additionally, by utilizing Navigation2, it enables navigation to the desired position and orientation within the experimental field. Fig.~\ref{fig:ic120ros2_architecture} shows the node configuration diagram of ic120\_ros2.

The roles of each node and the flow of operation commands in Fig.~\ref{fig:ic120ros2_architecture} are described below.

\begin{enumerate}
    \item First, the ic120\_com3\_ros converts the common control messages received from the IC120 into ROS 2 topic format. The values converted here include GNSS information and wheel odometry, which are used to estimate the position and orientation of the IC120.
    \item The GNSS information is first sent to gnss\_localizer\_ros2. The data includes the latitude, longitude, and altitude, which are position coordinates relative to the Earth-centered coordinate system. These coordinates need to be converted into position coordinates relative to the plane rectangular coordinate system, and this conversion is performed by gnss\_localizer\_ros2 before sending the data to ekf\_localization\_node.
    \item The ekf\_localization\_node receives the relative position information of the IC120 in the plane rectangular coordinate system from gnss\_localizer\_ros2, as well as the wheel odometry data from ic120\_com3\_ros, and uses this information to estimate the position and orientation of the IC120. The estimated position and orientation of the IC120 are then sent to Navigation2.
    \item Navigation2, using the position and orientation information of the IC120 received from ekf\_localization\_node, generates a path for the IC120 to navigate to the target location specified by the user and calculates the linear and angular velocities required for the machinery to follow that path. These linear and angular velocities are sent to ic120\_com3\_ros.
    \item Once ic120\_com3\_ros receives the linear and angular velocities in ROS 2 topic format from Navigation2, it converts them into common control messages in CAN format and sends them to the actual construction machinery, allowing it to operate.
\end{enumerate}

Additionally, when the ic120\_ros2 is launched, the current position and orientation are continuously sent from the machinery to Navigation2 node following steps 1, 2, and 3. This allows Navigation2 to continuously monitor the state of the IC120 and adjust its operations in real time as needed. As additional information, the DWB planner\cite{dwb_planner} is used for the local path planner, and the NavFn planner\cite{navfn_planner} is used for the global path planner.

\subsection{Manipulation module for backhoe ZX200}
\begin{figure}[htb]
\centering
\includegraphics[width=75mm]{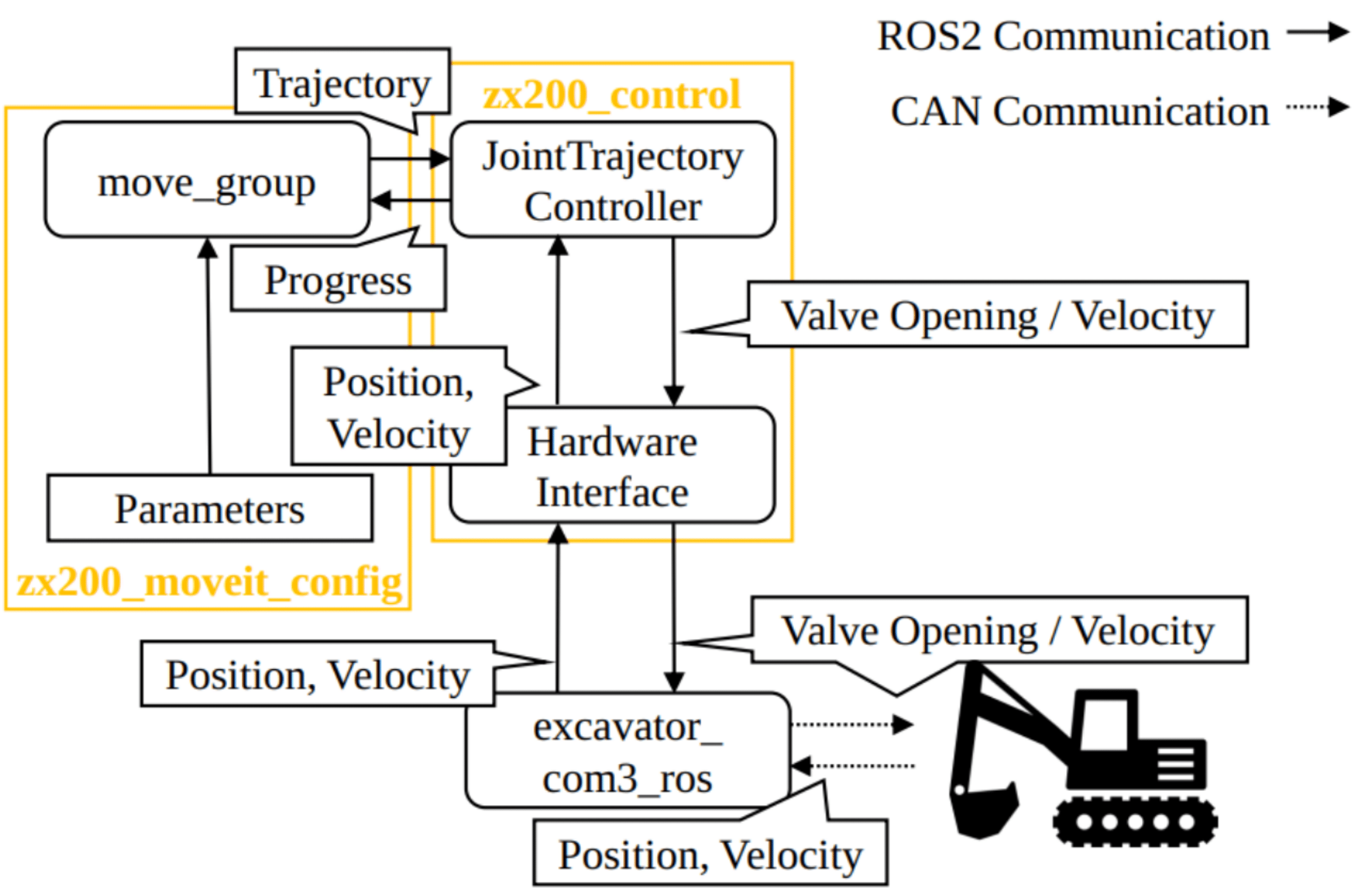}
\caption{The architecture of zx200\_ros2}
\label{fig:zx200ros2_architecture}
\end{figure}

For the manipulation control of the ZX200, we developed a set of ROS2 packages called zx200\_ros2~\cite{zx200_ros2}, which utilizes MoveIt!2. This allows for motion planning and execution using the swing, boom, arm, and bucket joints. By specifying the position and orientation of the end effector, the ZX200 can be moved to the desired posture. Fig.~\ref{fig:zx200ros2_architecture} shows the node configuration diagram of zx200\_ros2.

The roles of each node and the flow of operation commands in Fig.~\ref{fig:zx200ros2_architecture} are described below.

\begin{enumerate}
    \item First, zx200\_com3\_ros converts the common control messages received from the ZX200 into ROS 2 topic format, which include current joint position and angular velocity. 
    \item Then, by transmitting these data to the move\_group via the HardwareInterface and JointTrajectoryController, the move\_group can recognize the current posture of the ZX200.
    \item Then, the move\_group receives the target position and orientation of the end effector from user. Based on the current posture of the ZX200, it calculates the time-series trajectory of each joint angle (swing, boom, arm and bucket joints) required to achieve the target position and orientation of the end effector.
    \item The move\_group sends the calculated time-series trajectory to the JointTrajectoryController / HardwareInterface, which then calculates the valve openings and joint angular velocities required to achieve the desired posture for the hydraulically controlled construction machinery, and send this data to excavator\_com3\_ros.
    \item Once excavator\_com3\_ros receives the valve openings and joint angular velocity in ROS 2 topic format, it converts them into common control messages in CAN format and sends them to the actual construction machinery, allowing it to operate.
\end{enumerate}

Additionally, the state of the operating construction machinery (each joint angle and angular velocity) is constantly transmitted in the order of backhoe, excavator\_com3\_ros, HardwareInterface, JointTrajectoryController, move\_group. This allows MoveIt!2 to continuously monitor the state of the backhoe and flexibly adjust its operations in real time.

\section{ROS2-TMS for Construction modules}
\label{sec:modules}
\begin{figure}[htb]
\centering
\includegraphics[width=75mm]{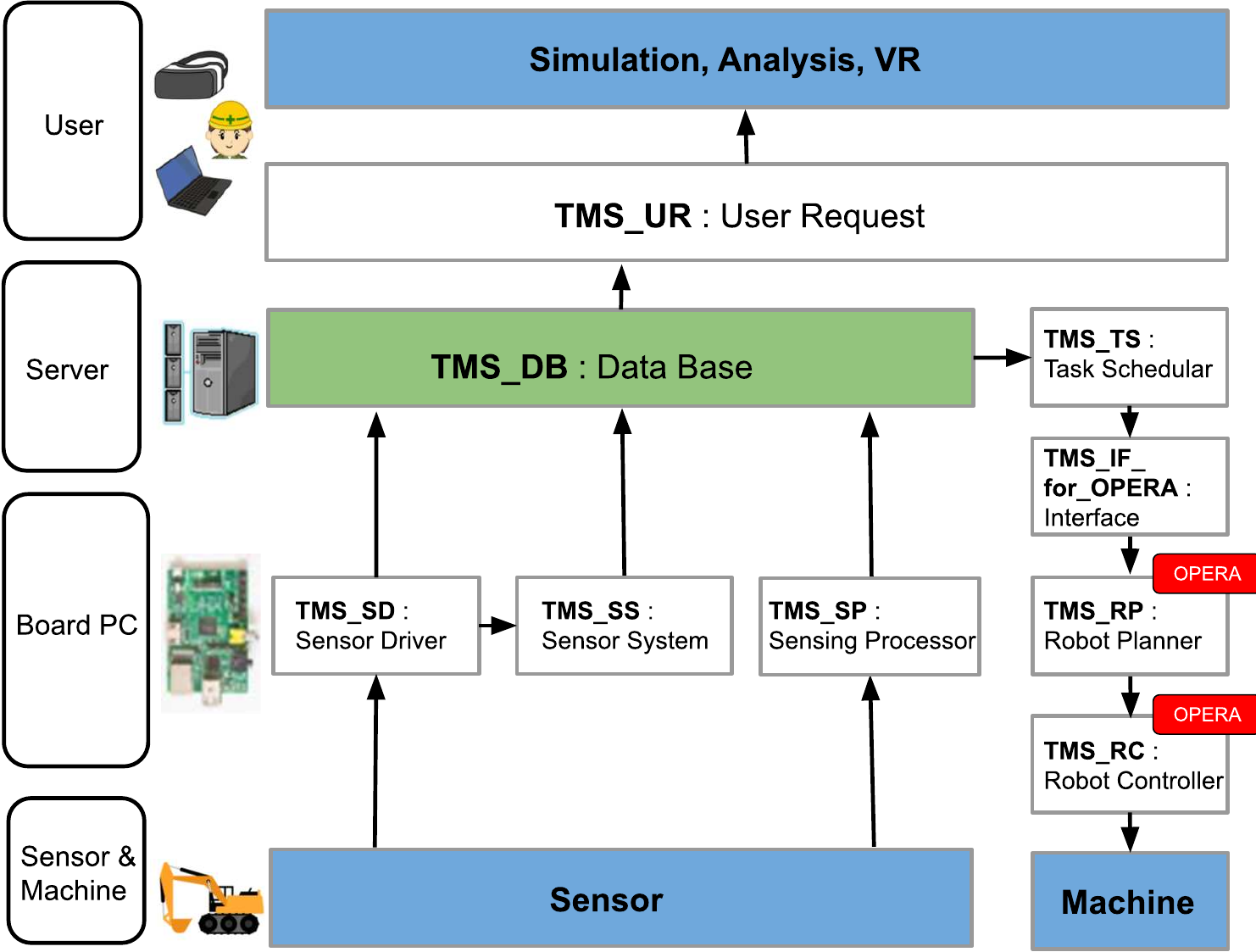}
\caption{The architecture of ROS2-TMS for Construction}
\label{fig:architecture_ros2-tms_for_construction}
\end{figure}

The overall architecture of ROS2-TMS for Construction is composed of various ROS2 modules, and these modules are connected according to the hierarchical structure shown in Fig.~\ref{fig:architecture_ros2-tms_for_construction}. 

An explanation of each module is provided below.

\begin{figure}[htb]
\centering
\includegraphics[width=75mm]{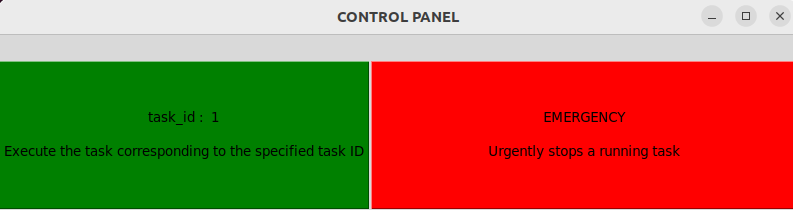}
\caption{The GUI button}
\label{fig:gui_button}
\end{figure}

\begin{itemize}
    \item \textit{DataBase Module} (TMS\_DB):\\ The module responsible for writing to and reading from the database (DB).   
    \item \textit{User Request Module} (TMS\_UR):\\ The module responsible for receiving requests from the user to ROS2-TMS for Construction. For example, one of the implemented features is the GUI button shown in Fig.~\ref{fig:gui_button}.
    \item \textit{Task Scheduler Module} (TMS\_TS):\\ The module that plays a given task using BT, which serves as a task scheduler of ROS2-TMS for Construction.
    \item \textit{Interface between ROS2-TMS for Construction and OPERA} (TMS\_IF\_for\_OPERA):\\ The module responsible for connecting ROS2-TMS for Construction to OPERA.
    \item \textit{Robot Planning Module} (TMS\_RP) [\textcolor{red}{OPERA}]:\\ The module responsible for manipulating and navigating construction machinery based on received commands from TMS\_TS.
    \item \textit{Robot Controller Module} (TMS\_RC) [\textcolor{red}{OPERA}]:\\ The module that calculates the signals for operating actual construction machinery (e.g., valve opening degree) based on the operation commands received in ROS2 topic format from TMS\_RP, converts them into common control messages in CAN format, and sends them to the actual construction machinery.
    \item \textit{Sensing Processor Module} (TMS\_SP):\\ The module that performs sensing processing on the collected sensor information (environmental data), and then sends the processed values to TMS\_DB. Based on this information, TMS\_DB updates the values in the DB in real time.
    \item \textit{Sensor Driver Module} (TMS\_SD): \\ The module responsible for sending data received from sensors directly to the TMS\_DB. It is used for handling data that does not require preprocessing before being stored in the database.
    \item \textit{Sensor System Module} (TMS\_SS): \\ The module responsible for storing large amounts of data received from sensors in its raw form in the DB via TMS\_DB.
    
\end{itemize}

The concept of ROS2-TMS for Construction is shown in Fig.~\ref{fig:ros_tms_for_construction}. Based on the concept of an informationally structured environment, data from sensors placed around the environment are stored in the database as "environmental information." Using this data, a real-time representation of the construction site is created in cyberspace, and the optimal actions are determined and corresponding commands are sent to the actual construction machinery. The former functionality for building a real-time construction site in cyberspace is referred to as the "OperaSimVR" while the latter functionality for operating actual construction machinery is called the "Task Management Mechanism." These are the two main features of ROS2-TMS for Construction, as described below.

\section{Task Management Mechanism}
As we transitioned ROS2-TMS\cite{ros2_tms} for use in construction sites, we overhauled the implementation, including the DB structure, while keeping the ROS2-TMS architecture. The major changes in this overhaul include switching the task scheduler from a Finite State Machine (FSM) to a Behavior Tree (BT), and integrating OPERA into some modules (TMS\_RP, TMS\_RC). This mechanism analyzes the given task using the Behavior Tree and employes OPERA to operate the actual construction machinery. 

\subsection{Task Execution flow}
\begin{figure}[htb]
\centering
\includegraphics[width=130mm]{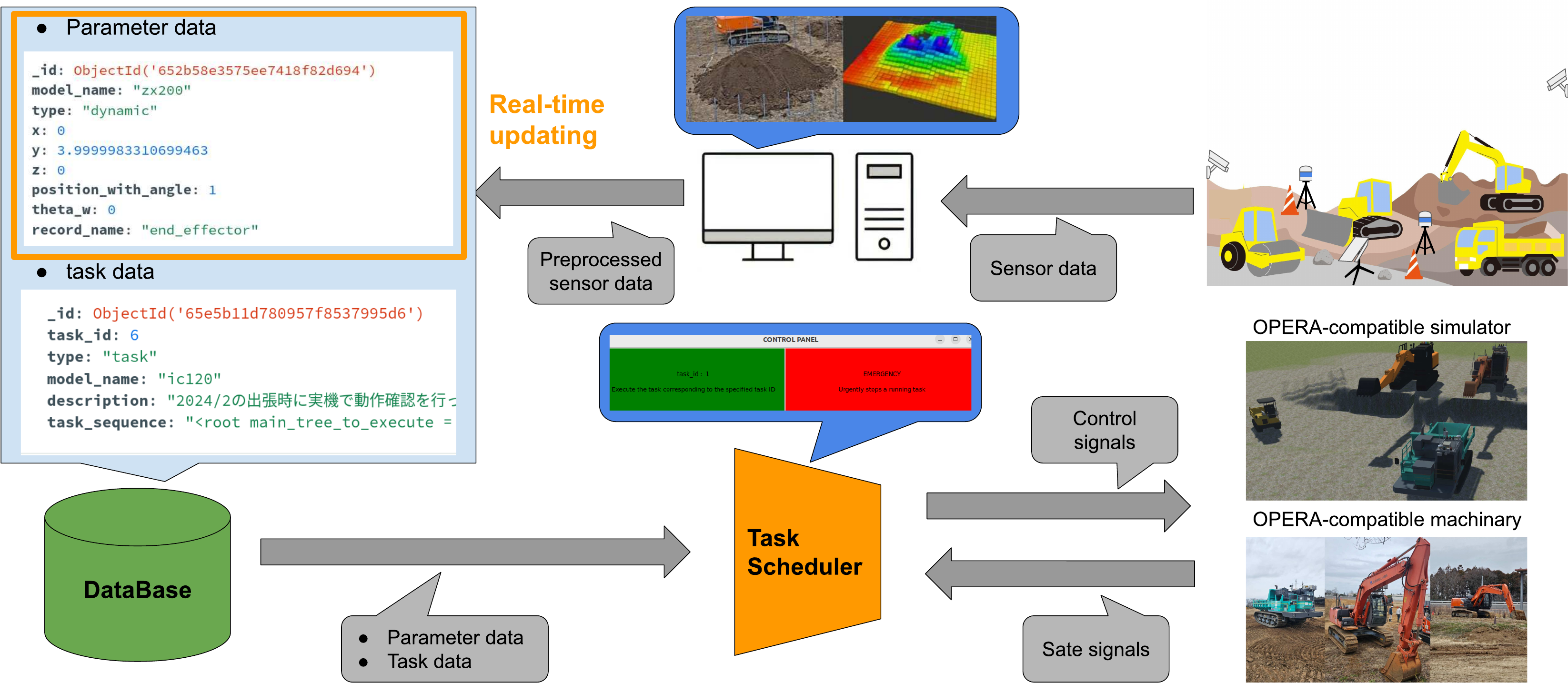}
\caption{The architecture of task management mechanism}
\label{fig:task_management_mechanism}
\end{figure}

The flow of construction operations executed by the task management mechanism is shown in Fig.~\ref{fig:task_management_mechanism}.

As shown in Fig.~\ref{fig:task_management_mechanism}, ROS2-TMS for Construction has a database-centered architecture. This database stores two types of data: "task data" and "parameter data." The details of each type of data are shown in Table~\ref{tbl:the_specification_of_task_data} and Table~\ref{tbl:the_specification_of_parameter_data}. Also, as shown in Table~\ref{tbl:the_types_of_parameter_data}, there are two types of parameter data: dynamic and static. Based on the contents of the DB and Fig.~\ref{fig:task_management_mechanism}, the flow of task execution using the task management mechanism can be described as follows:

\begin{table}[htb]
\vspace*{2mm}
 \caption{The specification of Task data}
 \label{tbl:the_specification_of_task_data}
 \centering
 \footnotesize
 \renewcommand{\arraystretch}{1.1}
 \begin{tabular}{c >{\centering\arraybackslash}p{8.5cm} c} 
   \hline
    Element & Description & Type \\\hline
    \_id & Automatically assigned by MongoDB & Required \\
    model\_name & Model name using task execution (e.g. ZX200) & Required \\
    description & Explanation of the task data & optional \\
    task\_id & Automatically assigned by ROS2-TMS for Construction & Required \\
    task\_sequence & Scenario data of task & Required \\\hline
 \end{tabular}
\end{table}

\begin{table}[htb]
\vspace*{2mm}
 \caption{The specification of parameter data}
 \label{tbl:the_specification_of_parameter_data}
 \centering
 \footnotesize
 \renewcommand{\arraystretch}{1.1}
 \begin{tabular}{c>{\centering\arraybackslash}p{9cm}c} 
   \hline
    Element & Description & Type \\\hline
    \_id & Automatically assigned by MongoDB & Repuired\\
    model\_name & Model name using task execution (e.g. ZX200) & Required \\
    type & Dynamic or static & Required \\
    record\_name & Name of parameter & Required \\
    others & Data used in the task processing (e.g. x,y,z) & Optional \\\hline
 \end{tabular}
\end{table}

\begin{table}[htb]
\vspace*{2mm}
 \caption{The static/dynamic parameter data}
 \label{tbl:the_types_of_parameter_data}
 \centering
 \footnotesize
 \renewcommand{\arraystretch}{1.1}
 \begin{tabular}{c >{\centering\arraybackslash}p{9.5cm}} 
   \hline
    Type & Description \\\hline
    Dynamic & Parameters dynamically updated during task execution by TMS\_SD, TMS\_SS, and TMS\_SP.\\
    Static & Parameters set before task execution and fixed during execution.\\\hline
 \end{tabular}
\end{table}

\begin{enumerate}
    \item First, the task execution command, including the user-specified task ID, is received by the user clicking the GUI button shown in Fig.~\ref{fig:gui_button}.
    \item The task management mechanism searches for the task data corresponding to the task\_id. If the data exists in the database, it retrieves the data and passes it to the task scheduler.
    \item The task scheduler analyzes the received task data, and once the analysis is complete, it checks the state of the construction machinery and sends the appropriate control signals to the construction machinery at the right timing.
    \item While the construction machinery is operating, its status is continuously sent to the task scheduler, which uses this information to determine the next action of the machinery. 
\end{enumerate}

In this way, by continuously monitoring the state of the actual machinery, the task scheduler can send control signals that match the current state of the construction machinery. On the other hand, the information obtained from sensors installed inside of the each construction machinery is often limited and insufficient for fully autonomous construction. Therefore, we developed a mechanism that incorporates data from external sensors during task execution shown at the top of Fig.~\ref{fig:task_management_mechanism}. This flow is described below. \\

\begin{enumerate}
    \item Environmental information is collected from sensors placed at the construction site.
    \item The collected sensor data is processed, and only the necessary information is calculated, updating the values in the DB in real time (in some cases, the data is written directly to the DB).
    \item During task execution, the task scheduler reads the updated parameter data from the DB at the required timing. 
\end{enumerate}

In this way, the task scheduler can operate adaptively by reading the parameters in the DB, which are updated in real time, at the desired timing, taking into account the information from sensors placed in the construction site.

\subsection{Task Scheduler}
\begin{figure}[htb]
\centering
\includegraphics[width=75mm]{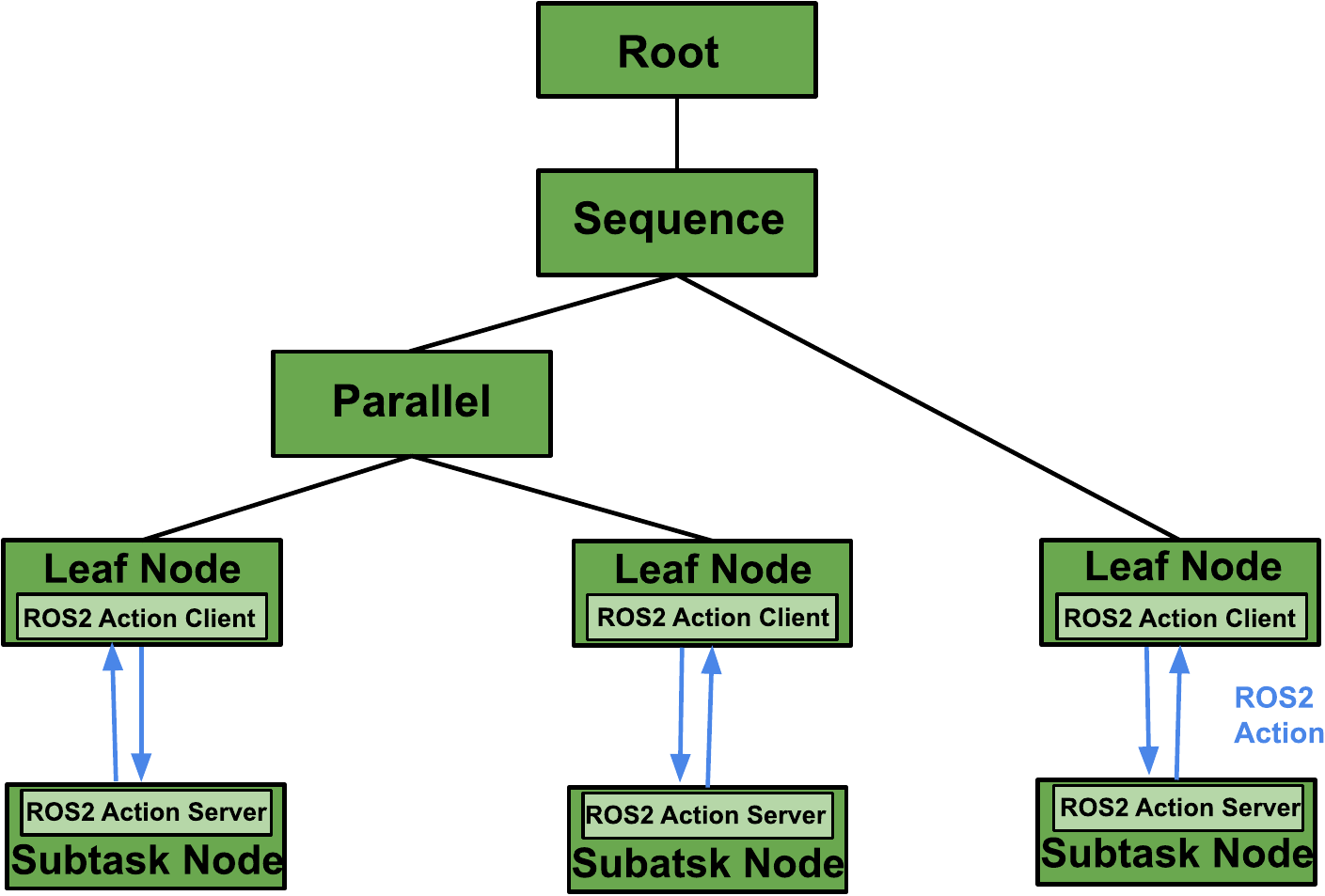}
\caption{The extended behavior tree}
\label{fig:customized_behaviortree}
\end{figure}

The task scheduler in the task management mechanism employs a Behavior Tree. Specifically, it uses a BehaviorTree.CPP\cite{behaviortree_cpp} with some extensions. Based on the task tree shown in Fig.~\ref{fig:customized_behaviortree}, the flow of task execution is explained.

In a Behavior Tree, as shown in Fig.~\ref{fig:customized_behaviortree}, a task is structured in a tree-like format. When a task is executed, the nodes are sequentially executed from the root, in accordance with the characteristics of each individual node. The key points in the task tree shown in Fig.~\ref{fig:customized_behaviortree} are the Leaf Nodes and Subtask Nodes. Their respective roles are divided as follows.

\begin{itemize}
    \item \textit{Leaf Node}: The node where three types of keys can be specified: "model\_name" and "record\_name," which are used to search the corresponding parameter data shown in Table~\ref{tbl:the_specification_of_parameter_data} and Table~\ref{tbl:the_types_of_parameter_data}, and "subtask\_name," which designates the Subtask Node to connect to. Additionally, it is possible to connect with the Subtask Node using ROS2 Action and receive the execution status of the subtask from the Subtask Node. By converting between ROS2 signals and BT signals, the status signals sent from the subtask can be transmitted to the Behavior Tree. This mechanism of the Leaf Node allows the Behavior Tree to manage the state of Subtask Nodes implemented in ROS2.
    
    \item \textit{Subtask Node}: This node is responsible for implementing subtask processing and cancel processing in ROS2 Action. When the Subtask Node is called, it calculates the target position and orientation that the construction machinery should achieve based on the values read from the DB, and sends these values to TMS\_RP or TMS\_RC as shown Fig.~\ref{fig:flow_of_task_execusion}. Additionally, if an emergency stop request is received via the Leaf Node from the Behavior Tree, the predefined camcel processing is executed to safely terminate the process.
\end{itemize}

    \begin{figure}[htb]
    \centering
    \includegraphics[width=100mm]{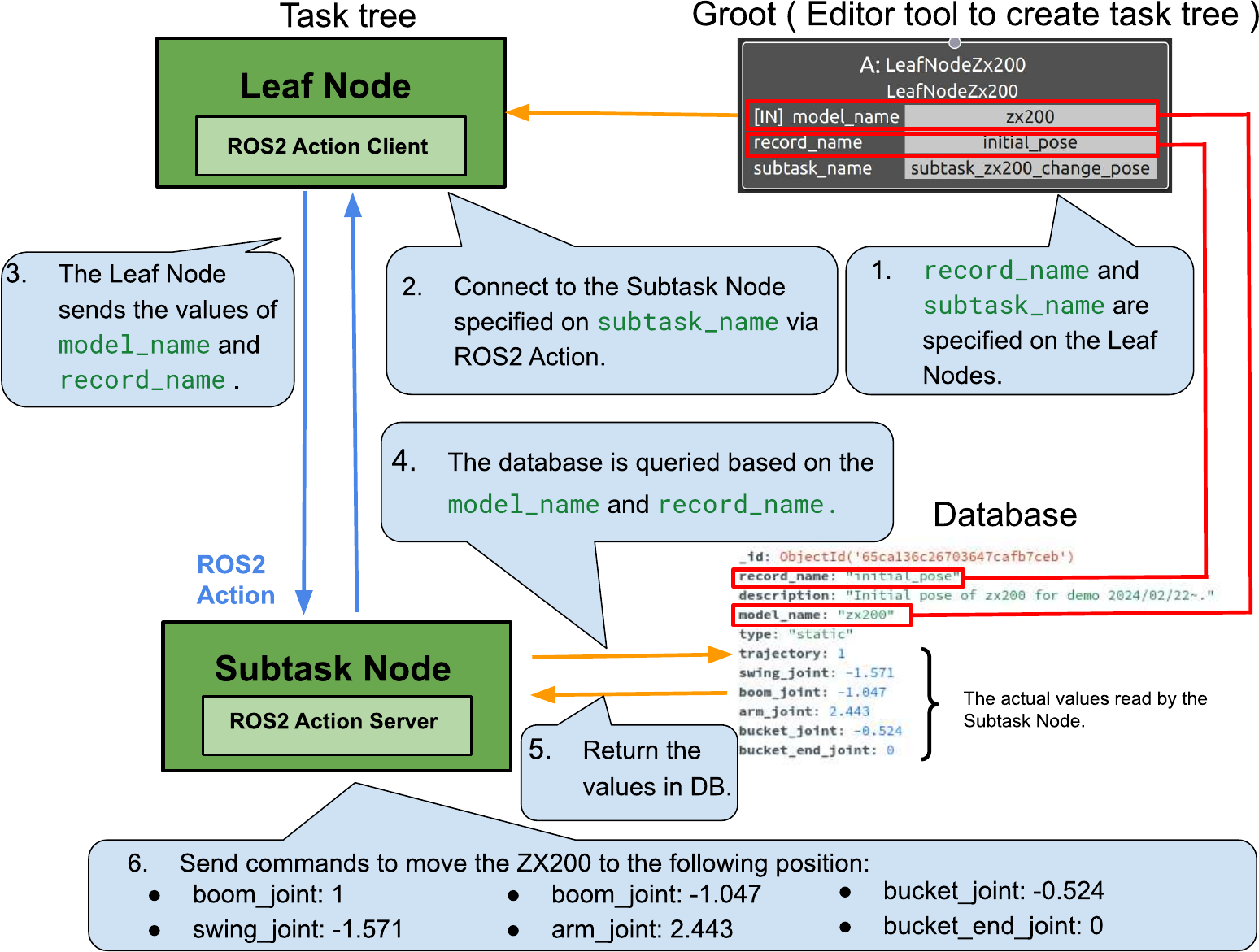}
    \caption{The flow diagram of the Leaf Node executing a Subtask Node.}
    \label{fig:LeafNode_and_SubtaskNode}
    \end{figure}

The flow of the Leaf Node executing a Subtask Node is shown in Fig.~\ref{fig:LeafNode_and_SubtaskNode}, and explained step by step. 

\begin{figure}[htb]
    \centering
    \includegraphics[width=50mm]{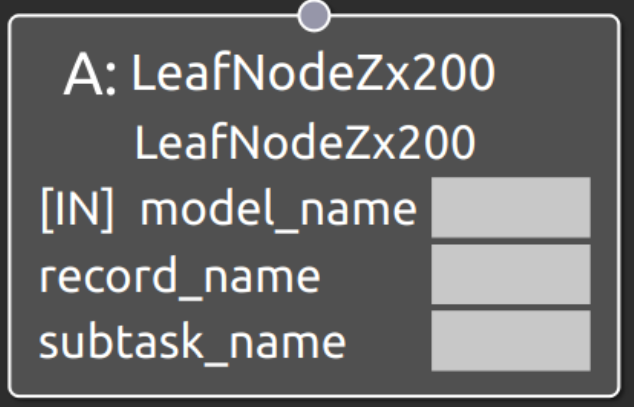}
    \caption{The Leaf Node shown on Groot.}
    \label{fig:LeafNode}
    \end{figure}

\begin{enumerate}
    \item As part of the preparation, when creating a task sequence, the values for three keys—model\_name, record\_name, and subtask\_name—are specified as parameters of the Leaf Node. (As described later, tasks can be implemented by connecting prepared nodes in a tree structure using a graphical tool called Groot. The user can directly input the parameter values on the Leaf Nodes, as shown in Fig.~\ref{fig:LeafNode}.)
    \item First, when the Leaf Node is called by task scheduler, it reads the three values specified in the task sequence: model\_name, record\_name, and subtask\_name. Then, it connects to the Subtask Node specified by subtask\_name via ROS2 Action.
    \item Once the connection between the Leaf Node and Subtask Node is established, the Leaf Node sends the values of the two keys—model name and record name—to the Subtask Node.
    \item When the Subtask Node receives the values of the two keys, it accesses the DB. It searches for the corresponding parameter data, and if match is found, it reads the values from the DB.
    \item The Subtask Node determines the appropriate action based on the read values. For example, as shown in Fig.~\ref{fig:LeafNode_and_SubtaskNode}, it sets the target posture using all the joint angles of the ZX200, which were read from the DB. The Subtask Node then requests motion planning and execution to TMS\_RP and TMS\_RC to achieve this target position and orientation shown in Fig.~\ref{fig:flow_of_task_execusion}.
    \item While the Subtask Node is executing the process, signals indicating its execution status are continuously sent from the Subtask Node to the Leaf Node.
\end{enumerate}

Both the Leaf Node and Subtask Node are reusable nodes that can be utilized multiple times within a single task. In the current implementation, separate Leaf Nodes are prepared for each construction machinery listed in Table~\ref{tbl:prepared_leaf_nodes}, and separate Subtask Nodes are prepared for each type of operation listed in Table~\ref{tbl:prepared_subtask_nodes}. Next, based on Fig.~\ref{fig:flow_of_task_execusion}, the process in which the Subtask Node operates ZX200 through TMS\_RP and TMS\_RC is explained as follows:

\begin{figure}[htb]
\centering
\includegraphics[width=135mm]{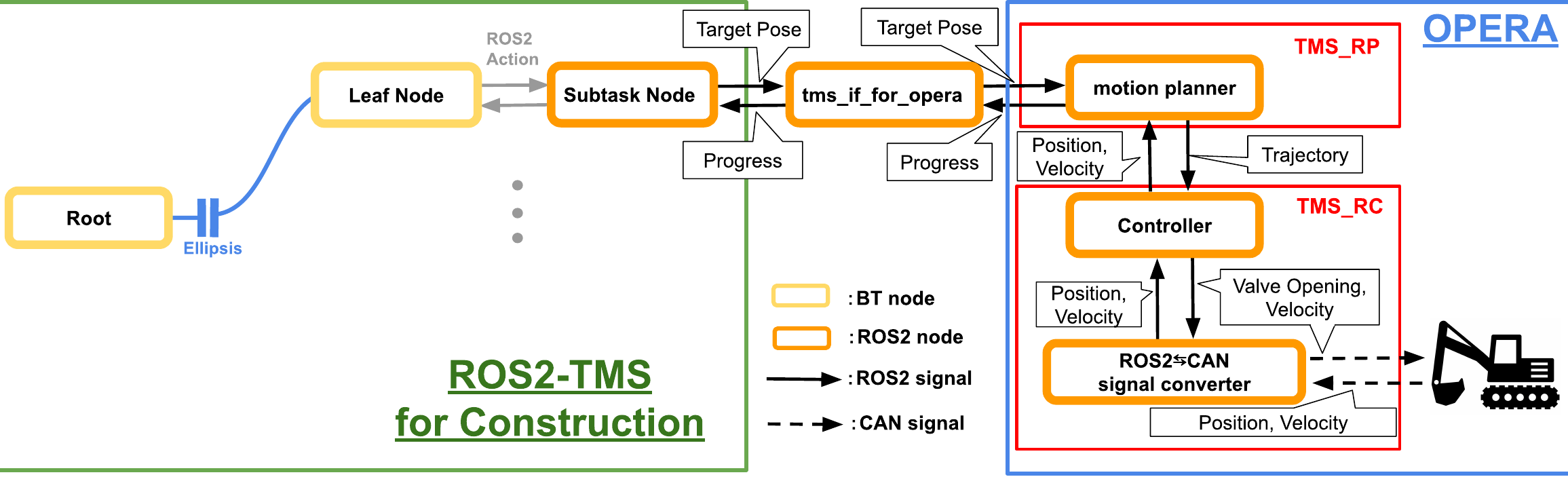}
\caption{The overall flow of task execusion of ZX200. The OPERA side is the same as shown in Fig.~\ref{fig:zx200ros2_architecture}, where the motion planner corresponds to move\_group, the controller corresponds to JointTrajectoryController/HardwareInterface and ROS2-CAN singnal converter correspond to excavater\_com3\_ros.}
\label{fig:flow_of_task_execusion}
\end{figure}

\begin{enumerate}
    \item After the task scheduler is activated by a user through the GUI button, the Leaf Node and Subtask Node are executed according to the task sequence shown in the flow of Fig.~\ref{fig:LeafNode_and_SubtaskNode}. Then, the target pose read by the Subtask Node are sent to the tms\_if\_opera node. 
    \item The tms\_if\_for\_opera node is used to connect ROS2-TMS for Construction with OPERA. Once the tms\_if\_for\_opera node receives the target pose, it converts this data into a format that can be processed by MoveIt!2 implemented in OPERA and then sends it to the motion planner.
    \item After the motion planner receives this data, it checks the current pose of the construction machinery and calculates the time-series poses needed to achieve the target pose. The motion planner then sends the calculated time-series poses to the controller.
    \item After the controller receives time-series pose, it calculates the valve openings and joint angular velocities necessary to achieve the time-series pose for the hydraulically controlled construction machinery. Then, the controller sends the calculated data to the ROS2-CAN signal converter.
    \item After the ROS2-CAN signal converter receives this data, the node converts it from a ROS2 topic into common control messages in CAN format and sends it to the ZX200.
    \item Once the ZX200 receives the common control messages, it operates according to them. 
\end{enumerate}

Additionally, the position and angular velocity of each joint of ZX200 are returned in the order of the ZX200, ROS2-CAN signal converter, controller, and motion planner. The motion planner then returns one of the status signals (SUCCESS, FAILURE, or RUNNING) in the order of motion planner, tms\_if\_for\_opera, Subtask Node, Leaf Node, and Root Node. When the status signal reaches the Root Node, the task scheduler is able to recognize the execution status of the subtask. In this way, the task scheduler can monitor the progress of the subtask in real time. 

Fig.~\ref{fig:flow_of_task_execusion_ic120} shows the process in which the Subtask Node operates IC120 through TMS\_RP and TMS\_RC. The flow of Fig.~\ref{fig:flow_of_task_execusion} is as follows:

\begin{figure}[htb]
\centering
\includegraphics[width=135mm]{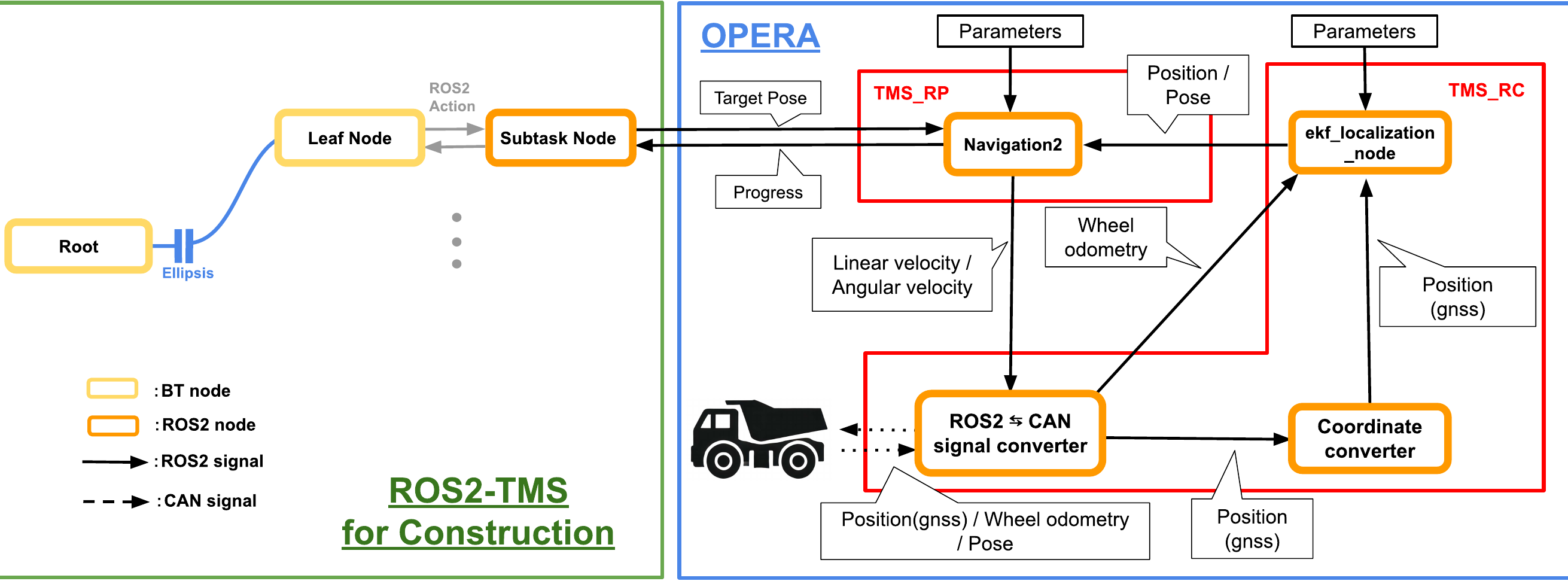}
\caption{The overall flow of task execusion of IC120. The OPERA side is the same as shown in Fig.~\ref{fig:ic120ros2_architecture}, where the Coordinate converter to gnss\_localizer\_ros2, and the the ROS2-CAN signal converter to ic120\_com3\_ros.}
\label{fig:flow_of_task_execusion_ic120}
\end{figure}

\begin{enumerate}
    \item First, the ROS2-CAN signal converter receives the common control messages in CAN format sent from the IC120 and converts them into ROS2 topic format. The data includes wheel odometry, GNSS position information, and joint angle information.
    \item The Coordinate Converter receives GNSS information from the ROS2-CAN signal converter and transforms this data, expressed in latitude, longitude, and altitude in the global coordinate system, into base coordinates in a plane rectangular coordinate system.
    \item The ekf\_localization\_node receives wheel odometry from the ROS2-CAN signal converter and position information in the plane rectangular coordinate system from the Coordinate Converter. These data are processed using an Extended Kalman Filter in ekf\_localization\_node. As the output of the Extended Kalman Filter, the position and orientation of the IC120 are calculated.
    \item After the task scheduler is activated by a user through the GUI button, the Leaf Node and Subtask Node are executed according to the task sequence read from the DB show in the flow of Fig.~\ref{fig:LeafNode_and_SubtaskNode}. The target position and orientation in IC120's navigation are send from the Subtask Node to the Navigation2.
    \item When the Navigation2 receives the target position and orientation for navigation, it retrieves IC120's current position and orientation from the ekf\_localization\_node.
    \item The Navigation2 calculates the linear velocity and angular velocity required for navigation and sends these values to the ROS2-CAN signal converter.
    \item Upon receiving the linear velocity and angular velocity, the ROS2-CAN signal converter converts them into common control messages in CAN format and transmits them to IC120.
\end{enumerate}

Additionally, the Navigation2 sends one of the values -SUCCESS, FAILURE, or RUNNING- as the navigation operation progress status to the Subtask Node. This status is propagated in sequence from the Navigation2 to the Subtask Node, Leaf Node, and then the Root Node. This allows the Behavior Tree to continuously monitor the progress of subtasks.

\subsection{Cancel mechanim}
\begin{figure}[htb]
\centering
\includegraphics[width=75mm]{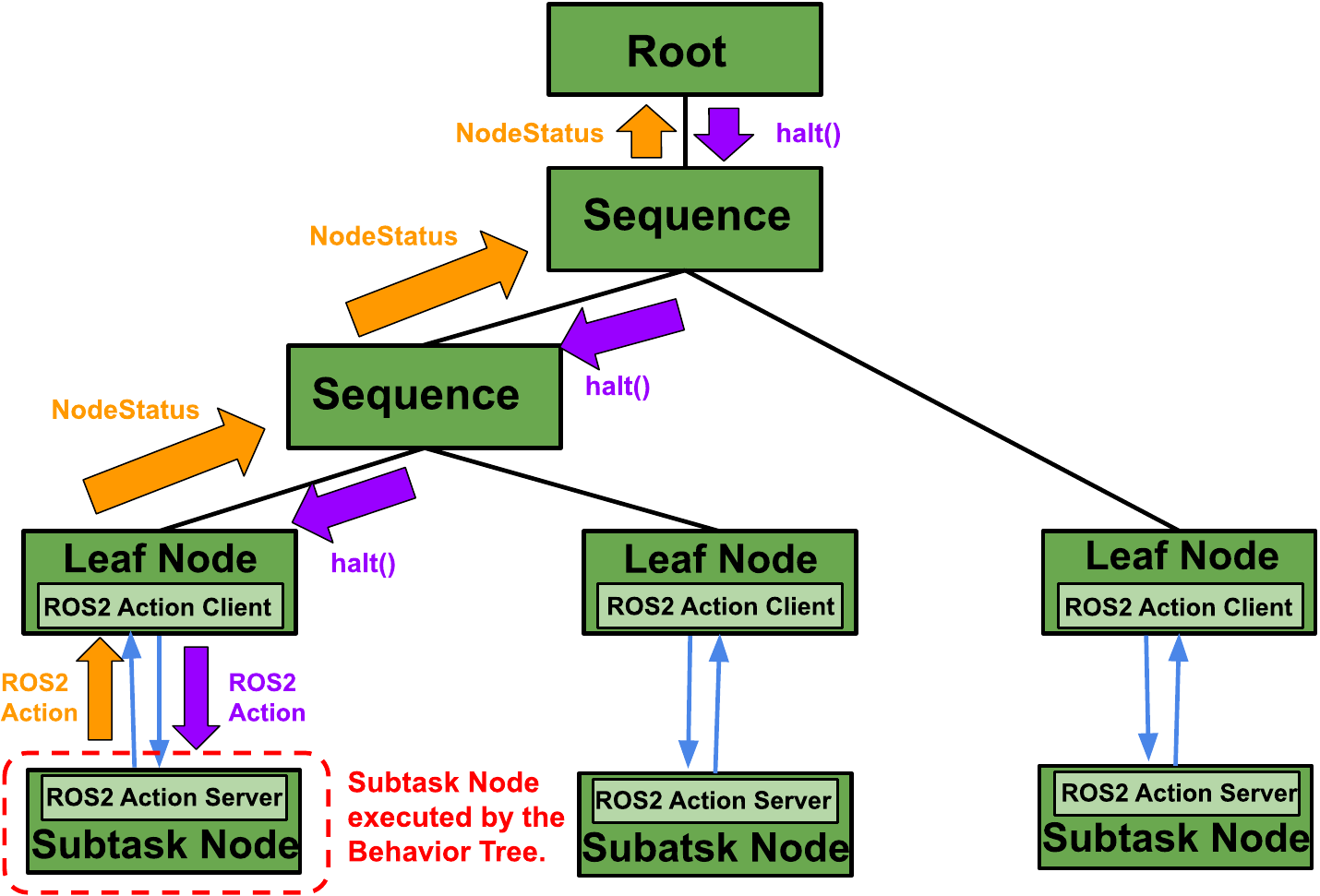}
\caption{The task cancel mechanism}
\label{fig:task_cancel_mechanism}
\end{figure}

The task management mechanism is equipped with a cancellation mechanism, allowing the user to safely terminate the entire task by clicking the emergency stop button shown in Fig.~\ref{fig:gui_button}. This ensures that the construction machinery has safely stopped before the task is fully terminated. An overview of this process is shown in Fig.~\ref{fig:task_cancel_mechanism}.

Explaining Fig.~\ref{fig:task_cancel_mechanism} step by step, it proceeds as follows:

\begin{enumerate}
    \item The task cancellation request is received from the user through the GUI button shown in Fig.~\ref{fig:gui_button}.
    \item Upon receiving the cancellation request, the task scheduler stops the ongoing task by having the parent node call the cancellation function of its child nodes sequentially, from the Root Node to the Leaf Node.
    \item When the cancellation function of the Leaf Node is called, a task cancellation request is sent to the connected Subtask Node through ROS2 Action.
    \item Upon receiving the task cancellation request from the Leaf Node, the Subtask Node interrupts the ongoing subtask process and performs the predefined cancellation operation.
    \item Once the Subtask Node completes the cancellation operation, it returns the result to the Leaf Node.
    \item The Leaf Node propagates the result back to the Root Node in reverse order, from the Leaf Node to the Root Node.
    \item Once the Root Node receives the result of the cancellation operation, the Behavior Tree finishes the entire task.
\end{enumerate}

In this way, the task management system is able to terminate the entire task after safely stopping the construction machinery.
\subsection{The method of task implementation}
\begin{figure}[htb]
\centering
\includegraphics[width=75mm]{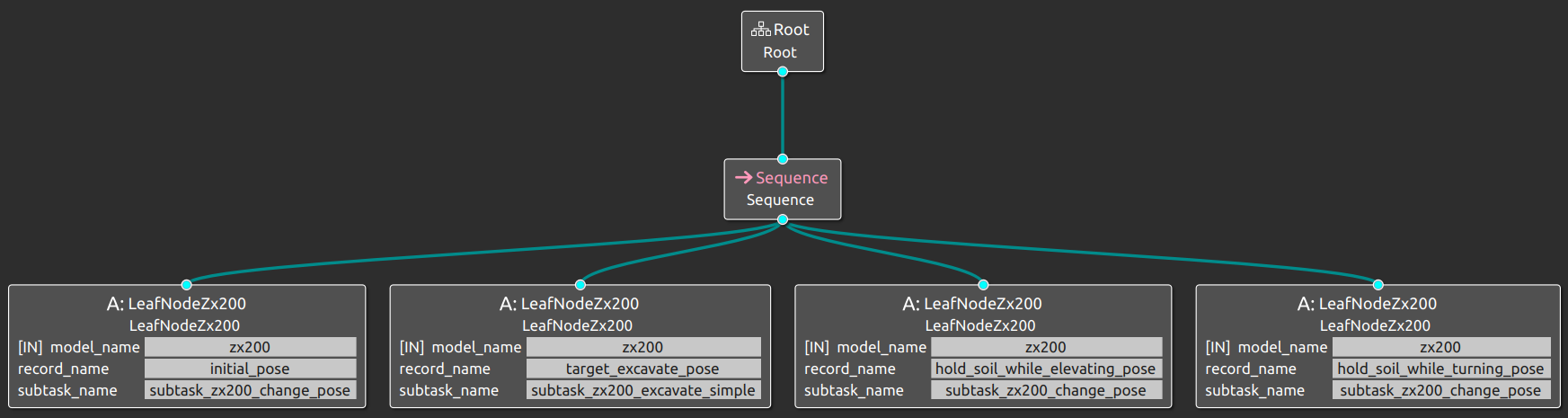}
\caption{The sample task tree}
\label{fig:sample_task_tree}
\end{figure}

In a Behavior Tree, a graphical tool called Groot is available for designing and implementing task sequences. Users can build a series of construction operations (tasks) by combining standard nodes provided by Groot with custom nodes provided by the ROS2-TMS for Construction. Fig.~\ref{fig:sample_task_tree} shows a sample task tree build on Groot.

\begin{figure}[htb]
\centering
\includegraphics[width=75mm]{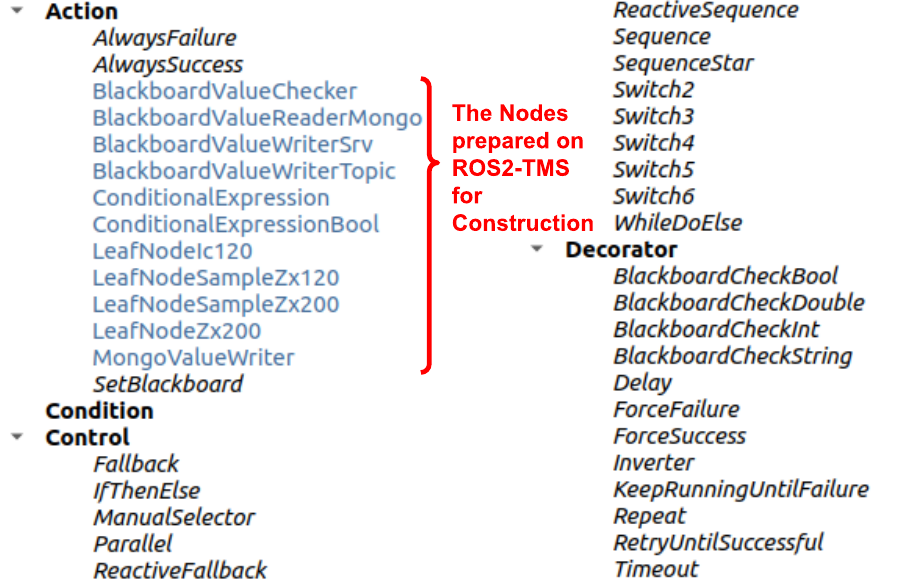}
\caption{The Nodes for task implementation}
\label{fig:nodes}
\end{figure}

\begin{figure}[htb]
\centering
\includegraphics[width=50mm]{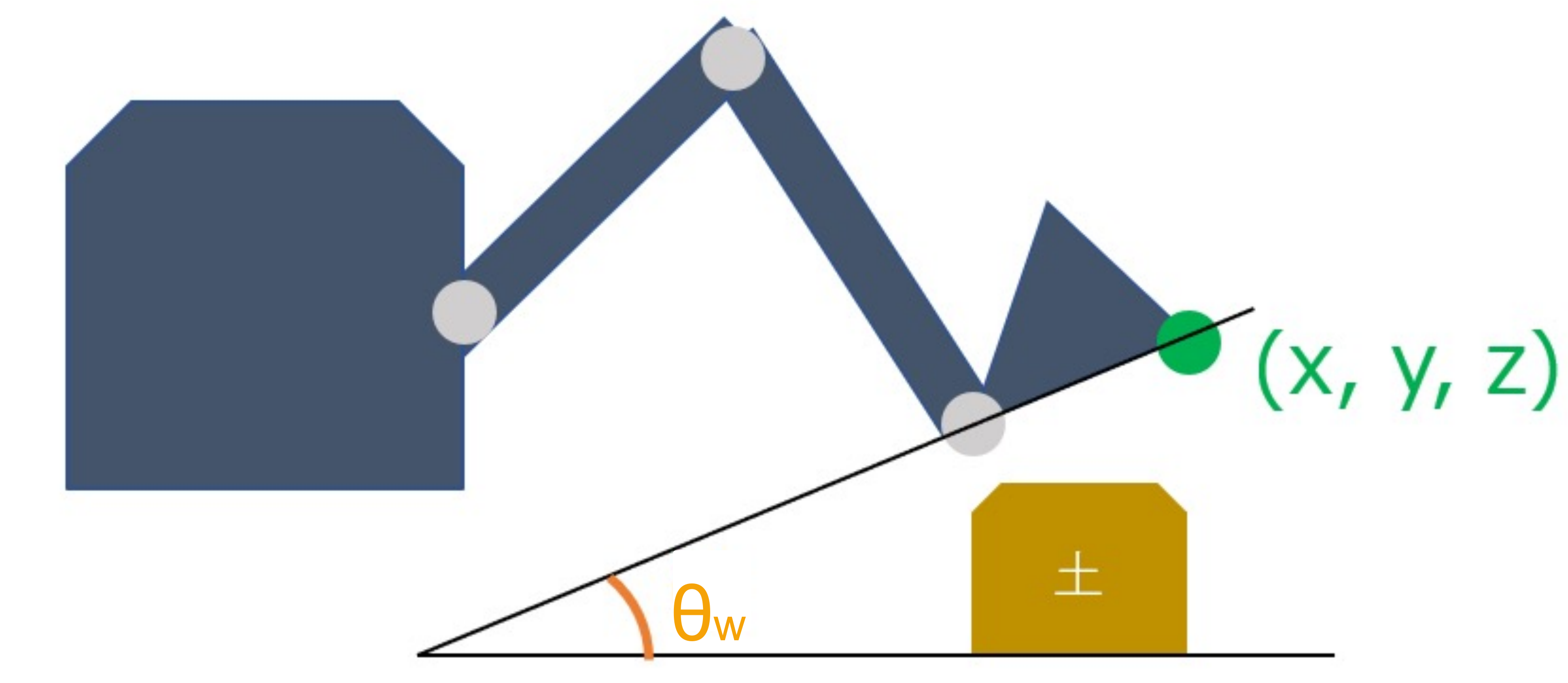}
\caption{The explanation of parameter "theta\_w"}
\label{fig:theta_w}
\end{figure}

\begin{figure}[htb]
\centering
\includegraphics[width=50mm]{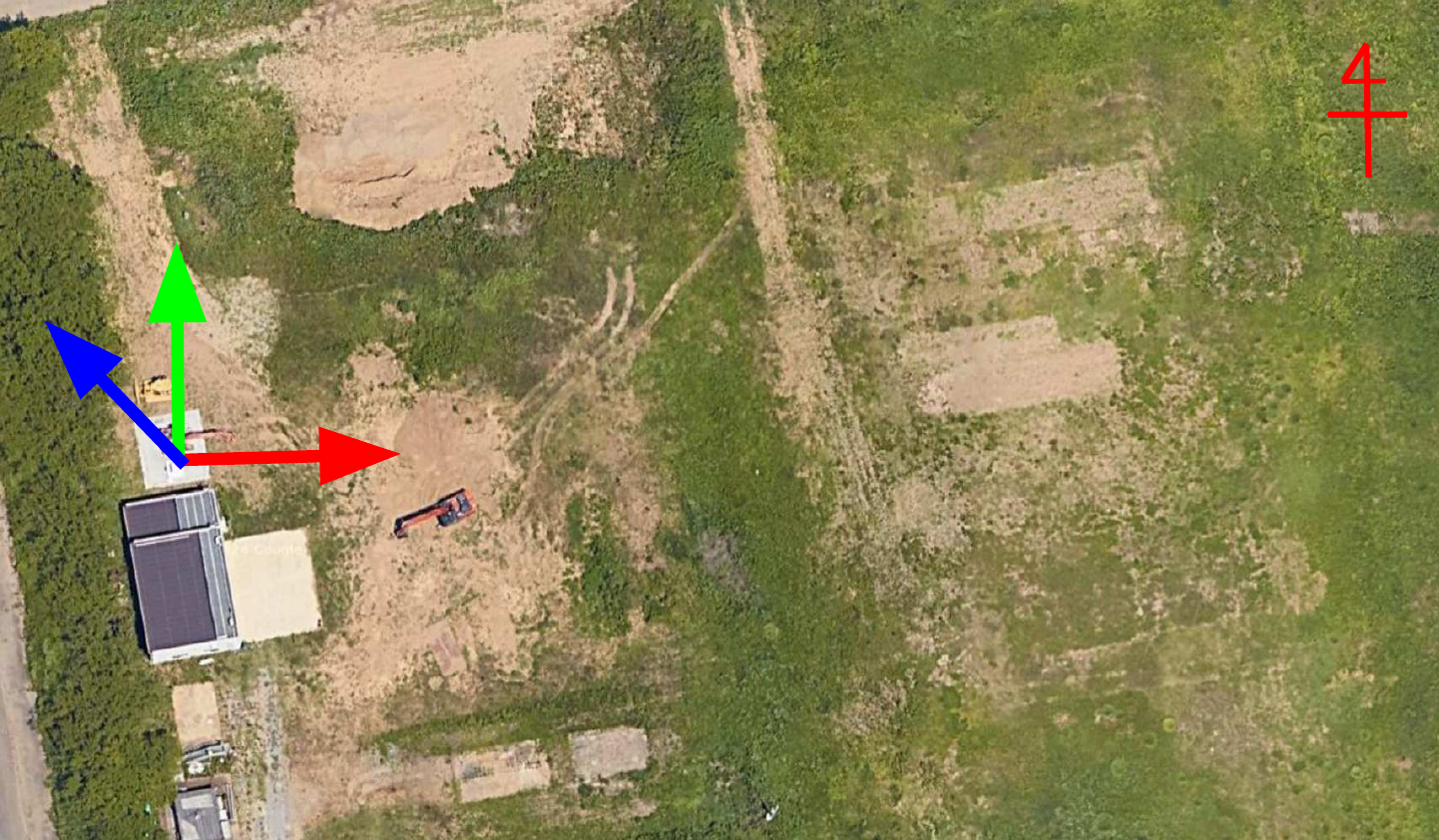}
\caption{The reference coordinate system placed within the PWRI experimental field is defined as follows: the x-axis is represented by red, the y-axis by green, and the z-axis by blue.}
\label{fig:map}
\end{figure}

As shown in Fig.~\ref{fig:nodes}, various nodes are prepared on Groot. By combining these nodes, users can easily implement from simple to complex operations, such as sequential processing, parallel processing, and parallel synchronized processing. In this way, users can design a series of construction operations without directly interacting with the program. As mentioned in the previous section, the behavior of the Subtask Node can be modified by changing the parameter values of the Leaf Node on Groot. Therefore, users can implement various processes by utilizing the values in the DB and Groot. Additionally, The available Lead Nodes and Subtask Nodes that the user can utilize when creating the task tree are shown in Table~\ref{tbl:prepared_leaf_nodes}, Table~\ref{tbl:prepared_subtask_nodes}, respectively. Here, theta\_w in Table~\ref{tbl:prepared_subtask_nodes} represents the angle between the bucket of ZX200 and the ground, as shown in Fig.~\ref{fig:theta_w}. The parameters qx, qy, qz, and qw denote the relative orientation to the reference map coordinates in quaternion form. The parameter yaw represents the relative orientation, with 0 degrees set along the x-axis direction in the map coordinates. The parameters x, y, and z indicate the relative position from the map coordinates, measured in meters. In the experimental field at PWRI, local coordinates are roughly set as shown in Fig.~\ref{fig:map}, with the east direction aligned with the x-axis and the north direction with the y-axis. During the navigation of the IC120, coordinate specification is based on this reference.

\begin{table}[htb]
\vspace*{2mm}
 \caption{Prepared Leaf Nodes}
 \label{tbl:prepared_leaf_nodes}
 \centering
 \footnotesize
 \renewcommand{\arraystretch}{1.1}
 \begin{tabular}{c c>{\centering\arraybackslash}p{8cm}} 
   \hline
    Machinery & Leaf Node & Description\\\hline
    IC120 & LeafNodeIc120 & Used when connecting the Subtask Node for IC120 operations to the task tree.\\
    ZX200 & LeafNodeZx200 & Used when connecting the Subtask Node for ZX200 operations to the task tree.\\\hline
 \end{tabular}
\end{table}

\begin{table}[p]
\vspace*{2mm}
 \caption{Examples of the Subtask Nodes}
 \label{tbl:prepared_subtask_nodes}
 \centering
 \footnotesize
 \renewcommand{\arraystretch}{1.1}
 \begin{tabular}{>{\centering\arraybackslash}p{1.2cm} >{\centering\arraybackslash}p{3cm} >{\centering\arraybackslash}p{3.5cm} >{\centering\arraybackslash}p{4cm}} 
   \hline
    Machine & Subtask Node & Parameters on DB & Description\\\hline
    IC120 & subtask\_ic120\_ follow\_waypoints & positions \& quaternions & Move IC120 from the starting point to the destination following waypoints.\\
    IC120 & subtask\_ic120\_ anyware & position \& quaternion & Move IC120 to the destination. \\
    IC120 & subtask\_ic120\_ navigate\_ through\_poses & positions \& quaternion	& Move IC120 from the starting point to the destination following waypoints, with orientation considered only at the final waypoint.\\
    IC120 & subtask\_ic120\_ release\_soil & target\_angle	& Control the vessel's opening and closing by specifying the opening angle. \\
    ZX200 & subtask\_zx200\_ change\_pose & 	position \& quaternion or positions \& yaw or all joint angles	& Move ZX200 to the target pose specified by end-effector's position or joint angles.\\
    ZX200 & subtask\_zx200\_ excavate\_simple & postions \& theta\_w & Execute the excavation operation for ZX200 by specifying the target excavation pose of the bucket. \\
    ZX200 & subtask\_zx200\_ release\_simple & target\_angle	& Control the opening and closing operation of the bucket of ZX200 by specifying the vessel joint angle.\\\hline
 \end{tabular}
\end{table}

\subsection{Task implementation policy}
In this study, a task refers to a construction operation executed by a task scheduler in one task execution cycle. The simplest task implementation in autonomous construction using multiple machinery is to operate several machinery from a single task. In this implementation, suppose there is a request to introduce new construction machinery during the execution of a task. In this case, ongoing construction task must either be stopped and execute a new task that can accommodate this scenario, or all possible scenarios must be incorporated into the task tree in advance. In the former case, even construction machinery that is unaffected by the addition would need to be stopped, resulting in decreased efficiency. In the latter case, the task tree would become large and complex. We found that such an implementation lacks flexibility. Therefore, we decided to limit the number of construction machines to be controlled to one per task, and to achieve a series of operations by coordinating multiple tasks. In addition to the situations mentioned above, it becomes possible to handle various other cases as well, as described below.\\

\begin{itemize}
    \item Since each construction machinery operates based on different tasks, if one of the construction machinery breaks down, only the task that operates the machinery and the task that could cause an accident, needs to be stopped, while the other tasks can continue to operate as they are.
    \item If an issue arises during the operation of construction machinery that is difficult to handle with autonomous construction, it is possible to stop only specific tasks and switch to manual operation for some of the machinery.  Additionally, once the difficult tasks are completed, it is also possible to resume work by executing tasks individually for each construction machine.
    \item If the work capacity becomes uneven at a construction site, it is possible to flexibly move construction machinery from areas with higher capacity to those with lower capacity through synchronization processing and switching the tasks executed by each machinery, as well as add the machinery to the work at those locations.  etc...
\end{itemize}

On the other hand, as the scalability of task design significantly improves, task design is generally expected to become more complex. Furthermore, in such a design, a framework for synchronizing individual tasks is essential. Therefore, we created a new information sharing mechanism called the global blackboard to handle synchronization between multiple tasks. The values in the global blackboard can be accessed by all tasks, and they can be used for synchronizing all ongoing tasks. Additionally, we added new Subtask Nodes to enable reading from and writing to the global blackboard from subtasks. Table~\ref{tbl:nodes_for_globalblackboard} shows the newly implemented Subtask Nodes for global blackboard. Note that BehaviorTree has a standard feature called a local blackboard. The local blackboard is a memory space that can only be referenced within each task and cannot be accessed by other tasks. The nodes provided by the behavior tree by default use the values stored in the local blackboard. Therefore, before using the values from the global blackboard, they must be read from the global blackboard, synchronized with the local blackboard, and then used in the subtask processing.

\begin{table}[htb]
\vspace*{2mm}
 \caption{Prepared Nodes for global blackboard}
 \label{tbl:nodes_for_globalblackboard}
 \centering
 \footnotesize
 \renewcommand{\arraystretch}{1.1}
 \begin{tabular}{c >{\centering\arraybackslash}p{9cm}} 
    \hline
    Subtask Name & Description\\\hline
    BlackboardValueChecker & Check local blackboard values and record the specified parameter when the subtask is executed.\\
    BlackboardValueReader & Read the global blackboard and store it on the local blackboard.\\
    MongoValueWriter & Read the local blackboard and store it on the global blackboard.\\\hline
 \end{tabular}
\end{table}

\section{Immersive VR Interface "OperaSimVR"}
We have been developing a mechanism to provide users with environmental information from sensors installed at earthwork sites and construction machinery. Our previously developed system\cite{ros2_tms_for_construction_vr} collected the position and orientation data of construction machinery at earthwork sites using multiple 3D LiDARs placed in the environment and reproduced the operation of the machinery in a simulator we developed. However, this system was unable to confirm the posture of the machinery’s arm or visualize operations such as excavation of soil mounds. Therefore, we developed a new system to more faithfully reproduce the earthwork site, including the arm movements of the machinery in cyberspace. This new system is based on OperaSim, the official simulator of OPERA using the physics engines PhysX and AGX on Unity. The system we developed receives environmental information such as terrain and the position and orientation of heavy machinery from a camera, GNSS, IMU, and angle sensors mounted on drones and heavy machinery. It then reproduces the earthwork site and the operations of the machinery in cyberspace. Construction machines in the real world and their models in cyberspace interactively work together in an immersive VR interface, which we name OperaSimVR. In the following sections, we report the visualization capabilities of OperaSimVR. The following are the elements that can be reproduced with this system.\\

\begin{itemize}
    \item The construction machinery \\ 
    Currently, it is possible to visualize OPERA-compatible construction machinery, including the backhoes (zx120 and zx200) and the crawler dumps (c30r and ic120), in cyberspace. 
    \item Terrain \\ 
    Using point cloud data from drones and other sources, it is possible to recreate the terrain by deforming the ground in the simulator using a physics engine such as PhysX or AGX. By applying textures such as aerial photographs to  the ground, the terrain can be realistically recreated.
    \item Buildings \\ By photographing the exterior of buildings and applying these images to objects, buildings can be recreated and placed on the terrain to match the actual site. 
\end{itemize}

\begin{figure}[htb]
\centering
\includegraphics[width=75mm]{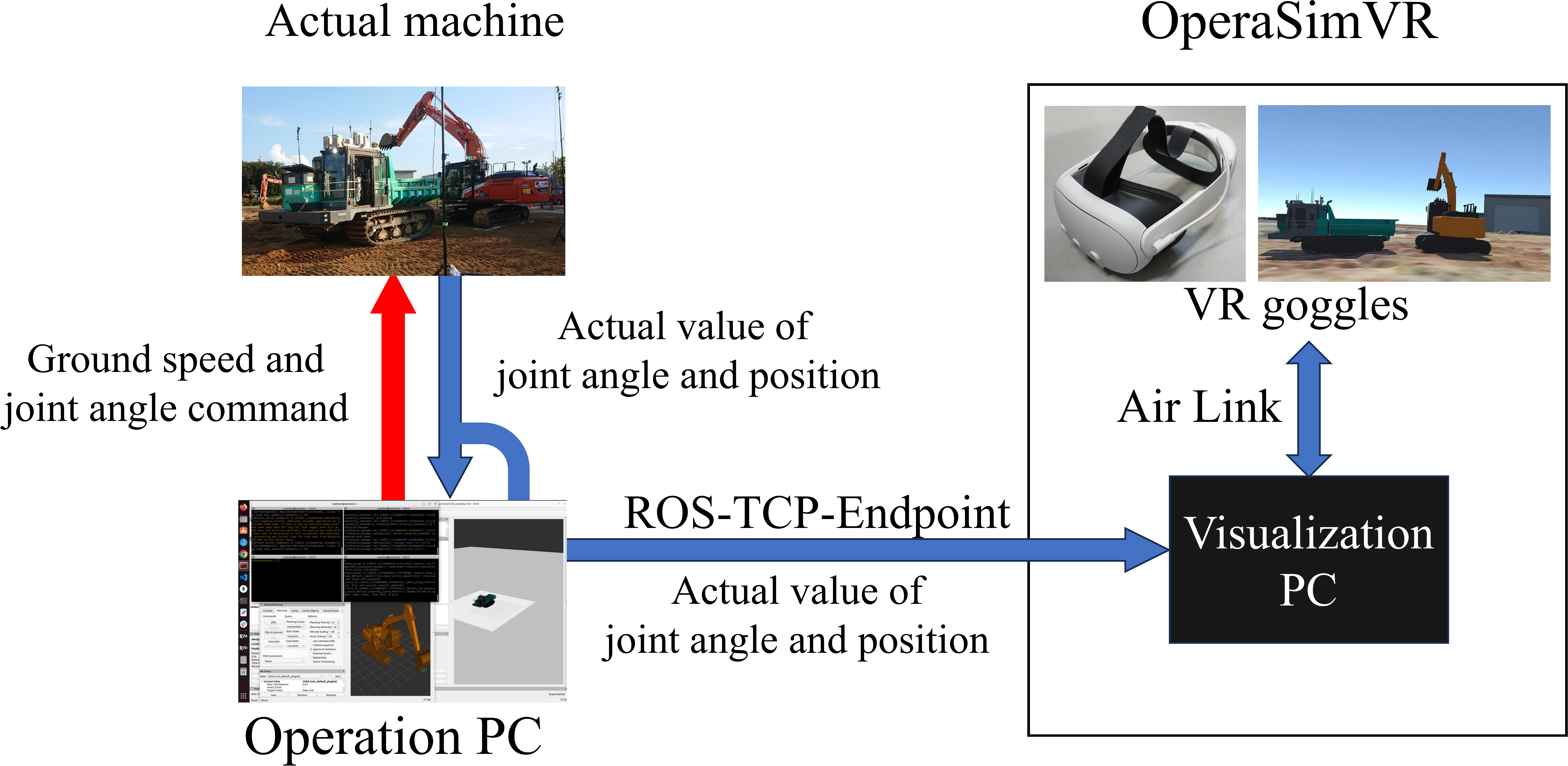}
\caption{Connection between actual machines, opearational PC, and OperaSimVR.}
\label{fig:vr_1}
\end{figure}

By reproducing these elements, the actual earthwork site, including heavy machinery, can be created in cyberspace. The following explains in detail the movements of the heavy machinery in the developed OperaSimVR. In this system, position data and joint angle data from OPERA-compatible heavy machinery are received as ROS2 topics, and the position and joint angles of the machinery models move in sync with the actual machines. We have extended OperaSim to allow the real-time visualization of machinery operations in cyberspace. Specifically, since OperaSim is originally a simulator for actual machinery, the heavy machinery models in OperaSim receive ROS2 topics, such as crawler rotation angles and arm joint angle commands, from an operational PC running ROS2 nodes, just like the actual machines. These models perform calculations for gravity and friction, and operate accordingly (standard mode). The system also sends position and joint angle information back to the operational PC as feedback, just like the actual machines. We added a new "replay mode", which the machinery models in OperaSim move based on the actual position and joint angle data obtained from ROS2 topics as truth data. Furthermore, by using VR assets for Unity, it is possible to immerse oneself in the VR experience and observe the actual machinery’s movements in real-time. In this system, the virtual camera is moved to the operator's seat of the heavy machinery, allowing users to view the scene from the perspective of sitting inside the machine in cyberspace through VR. Not only can users move around the earthwork site in cyberspace, but they can also freely observe the scene from a bird's-eye view. By using OperaSimVR, it is possible to observe areas that are difficult or dangerous to approach at earthwork site, such as near unstable ground prone to collapse, or poor visibility conditions like fog or night operations. Fig.~\ref{fig:vr_1} shows the flow of information transmission and reception when the actual machinery, operational PC, and OperaSimVR are connected. The operational PC for controlling the machinery executes the ROS-TCP-Endpoint along with the node for machine operation and connects to the PC running OperaSimVR. The system and VR headset are connected via Air Link, allowing information from the actual machinery to be output to the VR headset. If the system is used without connecting the VR headset, the earthwork site in cyberspace can still be viewed from an overhead perspective on the PC running OperaSimVR.

The heavy machinery models in the system obtain the following two types of information:

\begin{itemize}
    \item Position Information \\ The system receives the topic including the vehicle's center position and orientation in the global coordinate system defined in the earthwork site, which sent by the actual machinery to the operational PC. Based on this information, it updates the position and orientation of the model in OperaSimVR.
    \item Joint Angle Information \\ The system receives the topic icluding the joint angles of the construction machinery, which is sent by the actual machine to the operational PC, and updates the joint angles of the model accordingly. For the hydraulic excavator, the system uses the actual joint angle data to control the swing, boom, arm, and bucket of the model in OperaSimVR.
\end{itemize}

\section{Experiments}
We conducted verification experiments on the previously mentioned task management mechanism, immersive VR interface "OperaSimVR" and the new features of OPERA. The experiments were carried out using the experimental field at PWRI. In the experiment, we used the crawler dump IC120 and the backhoe ZX200. The two types of experiments-one with the task management mechanism and the other with OperaSimVR-will be explained in separate chapters.

\subsection{The experiment of task management mechanism}
\begin{figure}[htb]
\centering
\includegraphics[width=75mm]{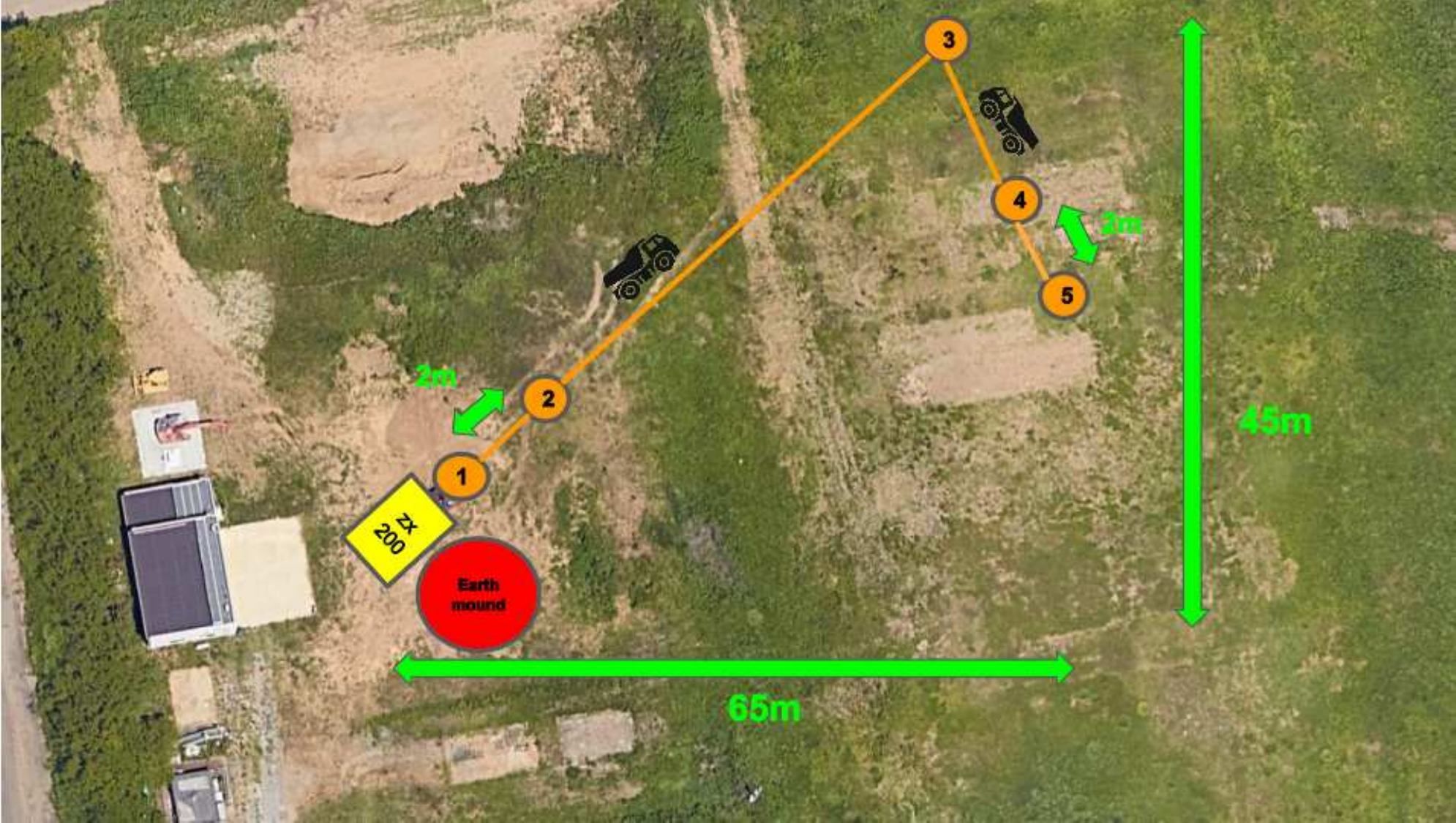}
\caption{The overview of the experimental field}
\label{fig:experimental_field}
\end{figure}

Fig.~\ref{fig:experimental_field} shows an overview of the experiments, indicating the initial points of the construction machinery and relay points in the field.

The details of each point in Fig.~\ref{fig:experimental_field} are as follows:

\begin{description}
    \item[Point 1] Position for loading soil onto the crawler dump from the mound using the backhoe ZX200.
    \item[Point 2] Position for aligning the orientation and reducing speed before arriving at the loading point (Point 1).
    \item[Point 3] Turning position for aligning the orientation towards the loading point (Point 1) and the dumping point (Point 5).
    \item[Point 4] Position for aligning the orientation and reducing speed before arriving at the dumping point (Point 5).
    \item[Point 5] Position for dumping soil from the crawler dump IC120.
\end{description}

In the initial state of the experiment, the IC120 was positioned at Point 2, and the ZX200 was facing the mound. The flowchart of the tasks used in this experiment is shown in Fig.~\ref{fig:task_flowchart}. The experiment was conducted according to the following steps:

\begin{figure}[p]
\centering
\includegraphics[width=135mm]{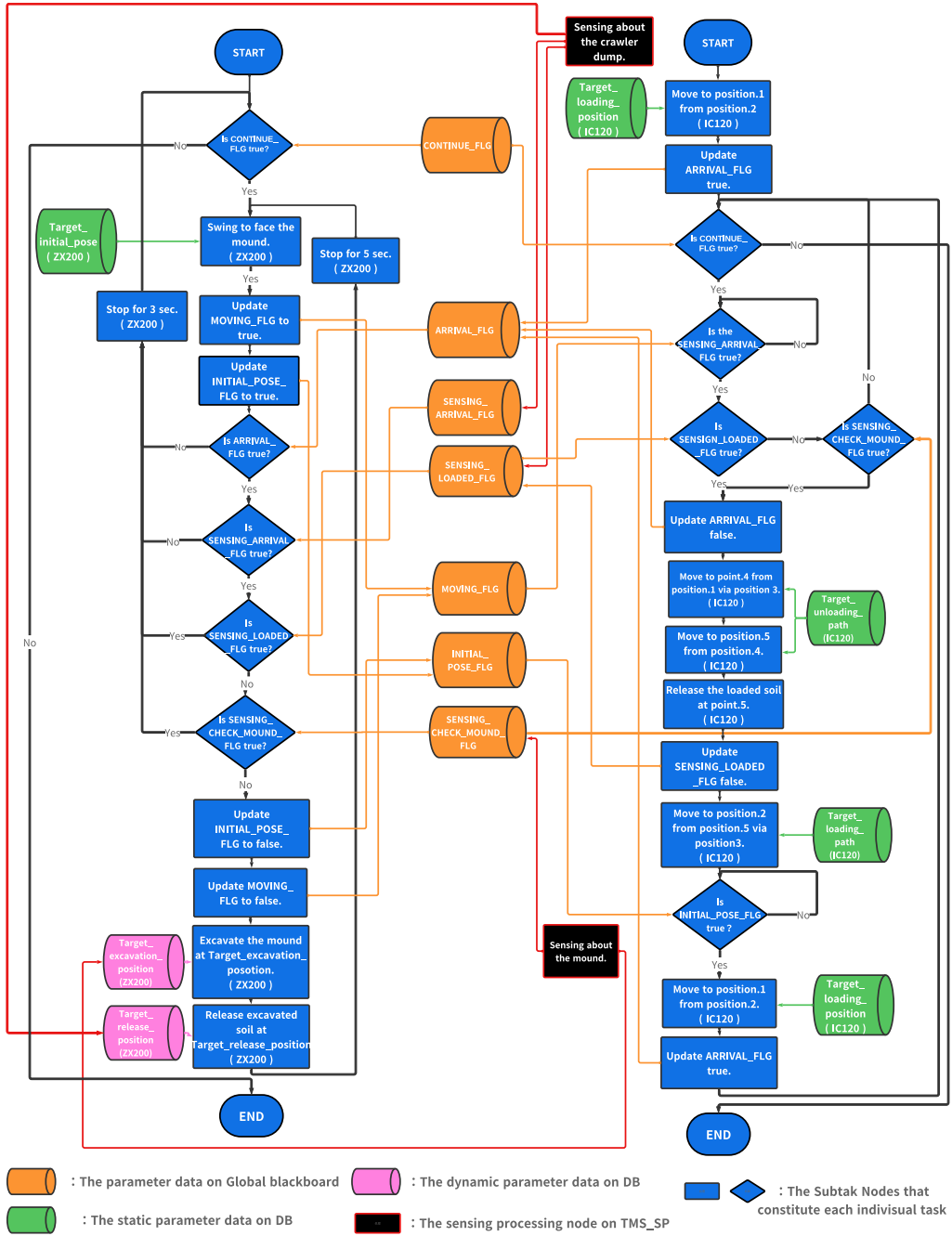}
\caption{The flowchart of the task (Left: ZX200, Right: IC120)}
\label{fig:task_flowchart}
\end{figure}

\begin{enumerate}
    \item First, when the task management mechanism receives a task execution command from the user via the GUI button, the tasks shown in Fig.~\ref{fig:task_flowchart} are started.
    \item IC120 first moves from Point 2 to Point 1, while the backhoe stops in front of the mound. 
    \item Once the IC120 has stopped at Point 1, the ZX200 begins the excavation operation at the mound and the loading operation onto the IC120. The excavation position and the loading position were dynamically updated based on the shape of the mound in cyberspace, which was built using sensor data from LiDAR and RGB cameras placed around the mound and the IC120.
    \item The ZX200’s excavating and loading operation repeats until either the IC120’s vessel is full of soil, or the soil in the mound is depleted to the point where excavation is no longer possible. Once either condition is met, the ZX200 stops, facing the front of the mound, and the IC120 moves to the dumping position (Point 5). 
    \item The IC120 travels from Point 3 to Point 4, then to Point 5, performing a turning maneuver at Point 3 to face its rear toward the unloading position (Point 5). At Point 4, it adjusts its orientation and decelerates toward the unloading position, then stops at Point 5 to perform the unloading operation.
    \item After the dumping operation at Point 5, the IC120 travels from Point 3 to Point 2, then to Point 1, performing a turning maneuver at Point 3 to face its rear toward the loading position (Point 1). At Point 2, it adjusts its orientation and decelerates toward the loading position, then stops at Point 1. 
    \item Check the flag on the global blackboard to determine whether the series of construction operations has been completed. If it has, terminate the entire process; if not, return to step.1.
\end{enumerate}

The details of the flags used on the global blackboard in the flowchart shown in Fig.~\ref{fig:task_flowchart} are provided in Table~\ref{tbl:parameters_on_globalblackboard}. Additionally, the list of dynamic parameters prepared for operating the ZX200 is shown in Table~\ref{tbl:dynamic_parameters}. As shown in Table~\ref{tbl:dynamic_parameters}, there are two methods for specifying the posture of the ZX200: one method specifies the angles of all joints, and the other specifies the position of the ZX200’s end-effector. In the latter method, "specifying the position of the ZX200’s end-effector," there are two approaches: one uses a quaternion to specify the position and orientation of the end-effector, and the other specifies it based on the angle theta\_w, which is the angle formed between the ground and the bucket joint, as shown in Fig.~\ref{fig:theta_w}. In this experiment, we used all three methods. Additionally, the estimation of the target excavation position based on the shape of the mound and the estimation of the target loading position based on the shape of the soil in the vessel of the IC120 were carried out using the sensing process. Specifically, in the sensing process, LiDAR and RGB cameras were placed around the mound and the IC120, and based on the information obtained from these sensors, the shape of the mound and the shape of the soil in the vessel were reconstructed in cyberspace. From this data, the target excavation position and the target dumping position were planned. These sensing values were written to the database, and it was confirmed that the task scheduler could read these values at the necessary timing as needed, allowing the actual construction machinery to excavate the target position on the mound and dump the soil at the target dumping position in the vessel. At the same time, we conducted experiments to confirm that the construction machinery in operation could be stopped using the task cancellation mechanism.

\begin{table}[htb]
\vspace*{2mm}
 \caption{Parameters on global blackboard}
 \label{tbl:parameters_on_globalblackboard}
 \centering
 \footnotesize
 \renewcommand{\arraystretch}{1.1}
 \begin{tabular}{>{\centering\arraybackslash}p{1.5cm} >{\centering\arraybackslash}p{2.5cm} c >{\centering\arraybackslash}p{2.5cm} >{\centering\arraybackslash}p{3cm} >{\centering\arraybackslash}p{1cm} } 
   \hline
    Update method & Parameter name & Type & IF True & IF False & Initial state\\\hline
    Auto & ARRIVAL\_ FLG & bool & The dump truck is at Position.1 (loading position). & The dump truck is not at Point.1 (loading position). & false\\
    Auto & CONTINUE\_ FLG & bool & Continuing the construction work. & Stopping the construction work. & true \\
    Auto & MOVING\_FLG & bool & Allow the dump truck to move from Point 1 (loading position). & Deny the dump truck to move from Point 1 (loading position). & true \\
    Auto & INITIAL\_ POSE\_FLG & bool & The ZX200 is facing the front of the mound. & The ZX200 isn't facing the front of the mound. & true \\
    Manual (Sensing) & SENSING\_ ARRIVAL\_ FLG & bool & The dump truck is at Point 1 (loading position). & The dump truck is not at Point 1 (loading position). & false\\
    Manual (Sensing) & SENSING\_ CHECK\_ MOUND\_FLG & bool & There is not enough soil in the mound. & There is enough soil in the mound. & false \\
    Manual (Sensing) & SENSING\_ LOADED\_FLG & bool & There is a sufficient amount of soil in the vessel. & There is not a sufficient amount of soil in the vessel. & false\\\hline
 \end{tabular}
\end{table}

\begin{table}[htb]
\vspace*{2mm}
 \caption{Dynamic Parameters}
 \label{tbl:dynamic_parameters}
 \centering
 \footnotesize
 \renewcommand{\arraystretch}{1.1}
 \begin{tabular}{>{\centering\arraybackslash}p{1.5cm} >{\centering\arraybackslash}p{4cm} >{\centering\arraybackslash}p{2.5cm} >{\centering\arraybackslash}p{4cm}} 
   \hline
    Update method & Parameter name & Type & Description \\\hline
    Fix & Target\_initial\_pose (ZX200) & all joint angles & The posture of the ZX200 when facing the front of the soil mound is defined. \\
    Fix & Target\_loading\_point (IC120) & position \& quaternion & The position and pose of IC120 at Position 1 is defined. \\
    Fix & Target\_loading\_path (IC120) & positions \& quaternion & The positions and poses of IC120 at waypoints (from Point 3 to Point 2) are defined. \\
    Fix & Target\_unloading\_path (IC120) & postions \& quaternions & The positions and poses of IC120 at waypoints (from Point 3 to Point 4 and Point 5) are defined. \\
    Manual (Sensing) & Target\_excavation\_position (ZX200) & postion \& theta\_w & The target excavation position of the soil mound by the ZX200 is defined.\\
    Manual (Sensing) & Target\_release\_position (ZX200) & position \& theta\_w & The target unloading position of the soil mound by the ZX200 is defined. \\\hline
 \end{tabular}
\end{table}

\subsection{The result of the experiment on task management mechanism}
\begin{figure}[htb]
\centering
\includegraphics[width=75mm]{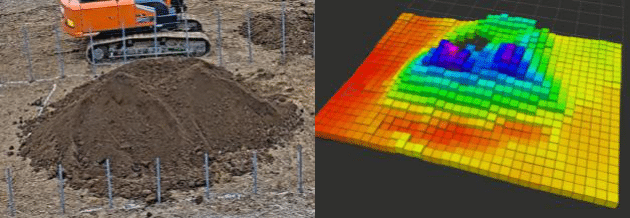}
\caption{Earth mound (left) and the map showing the distribution of earth mound heights (right) to estimate the recommended blade position for excavation.}
\label{fig:mound}
\end{figure}

\begin{figure}[htb]
\centering
\includegraphics[width=75mm]{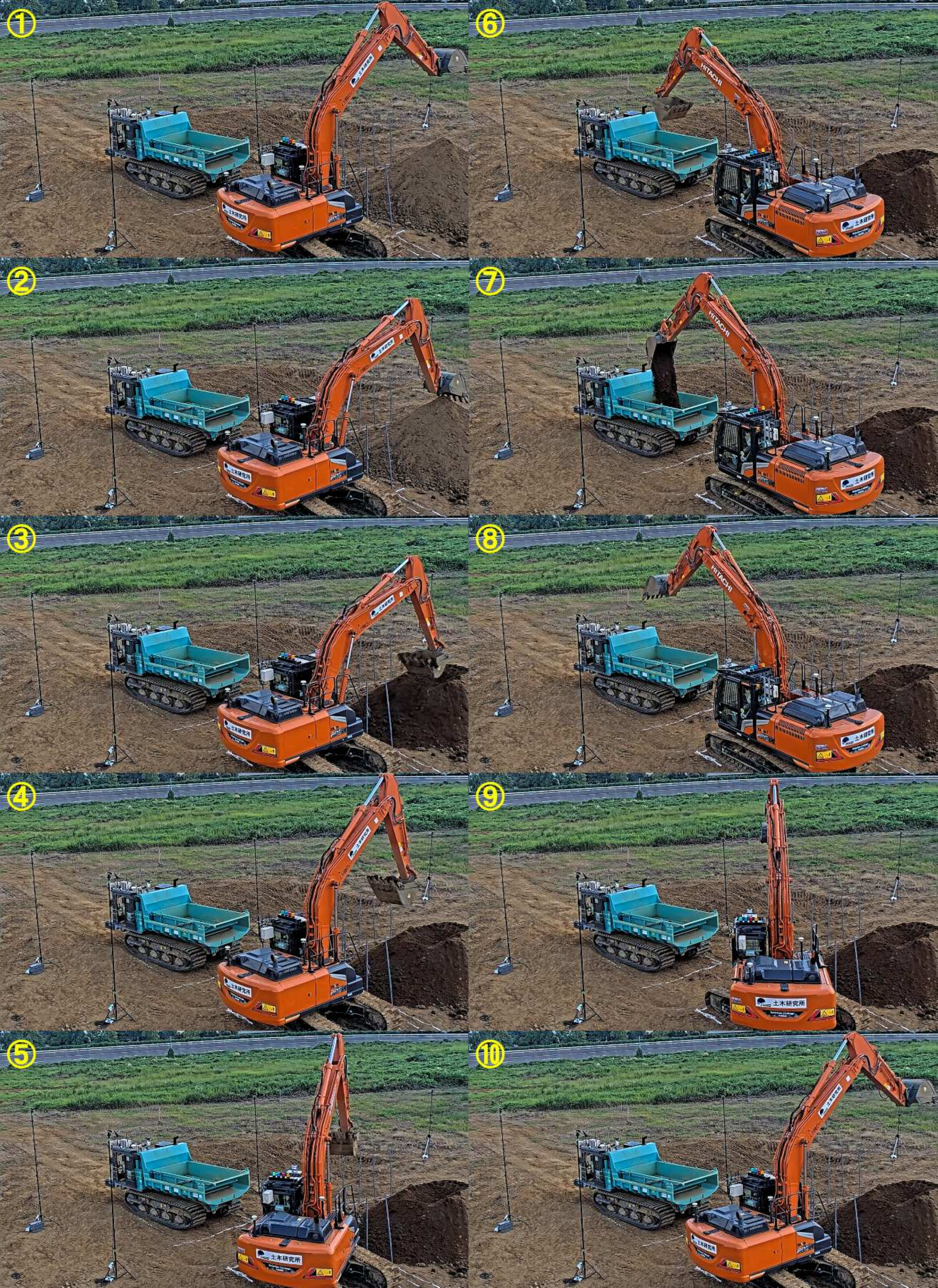}
\caption{Result of the task for backhoe}
\label{fig:experiment1}
\end{figure}

In Fig.~\ref{fig:experiment1}, following the previous flow, when the IC120 reached Point 1, the ARRIVAL\_FLG was automatically set to true. Then, the value of SENSING\_ARRIVAL\_FLG was changed from false to true via a ROS2 topic from the other PCs, and the ZX200 automatically began excavating soil. In this moment, the INITIAL\_POSE\_FLG on the global blackboard was automatically set to false. The backhoe repeatedly excavates soil from the mound and loads it onto the dump. In this process, the target excavation position on the mound, estimated from the mound constructed in cyberspace (as shown in Fig.~\ref{fig:mound}) based on sensor data obtained from LiDAR and RGB cameras placed around the mound and IC120, was actually excavated. Additionally, the soil loading operation into the IC120 was carried out based on the target dumping position estimated from the shape of the soil in the vessel of the IC120. Then, when SENSING\_LOADED\_FLG on the global blackboard became true or SENSING\_CHECK\_MOUND\_FLG became false, the backhoe stopped excavation and loading, faced the mound. Then, the task controlling the ZX200 automatically changes the INITIAL\_POSE\_FLG on the global blackboard to true. Following this, the task controlling the IC120 automatically changes the ARRIVAL\_FLG on the global blackboard to false, and the IC120 proceeds to Point 5 via Point 3 and Point 4 to dump the soil inside the vessel. Then, the task controlling the IC120 automatically changes the SENSING\_LOADED\_FLG on the global blackboard to false, and the IC120 proceeds to Point 1 via Point 3 and Point 2. Upon arrival, the task controlling the IC120 automatically changes the ARRIVAL\_FLG to true. It then checks the CONTINUE\_FLG on the global blackboard, and if it is false, the task for the IC120 is terminated. Meanwhile, the task for the ZX200 repeatedly checks the value of CONTINUE\_FLG while the IC120 is not at the loading position, so at this point, the task is considered complete. If the CONTINUE\_FLG on the global blackboard is true, the previous operations are repeated. 

\begin{figure}[htb]
\centering
\includegraphics[width=75mm]{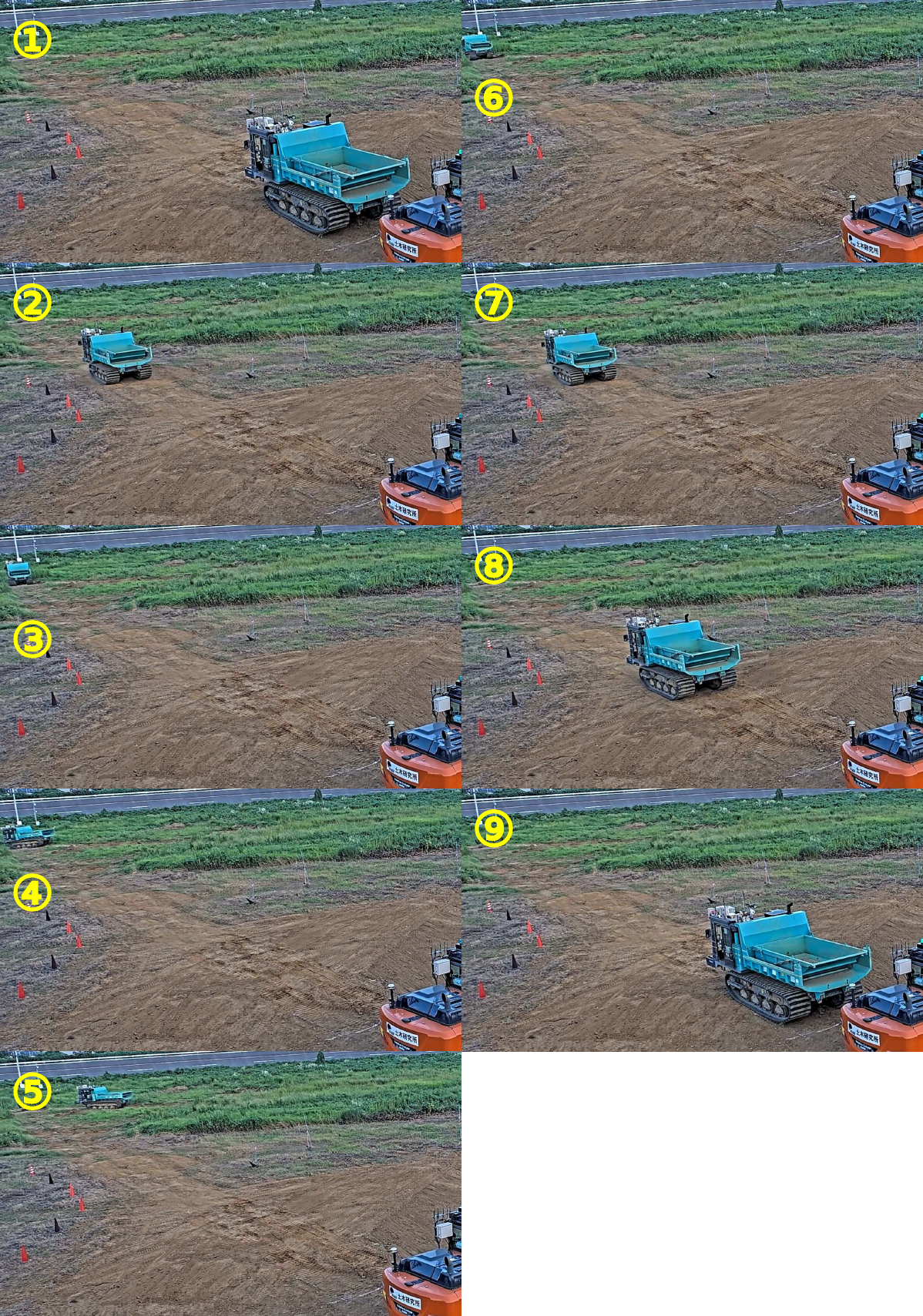}
\caption{Result of the task for crawler dump}
\label{fig:experiment2}
\end{figure}

\begin{figure}[htb]
\centering
\includegraphics[width=75mm]{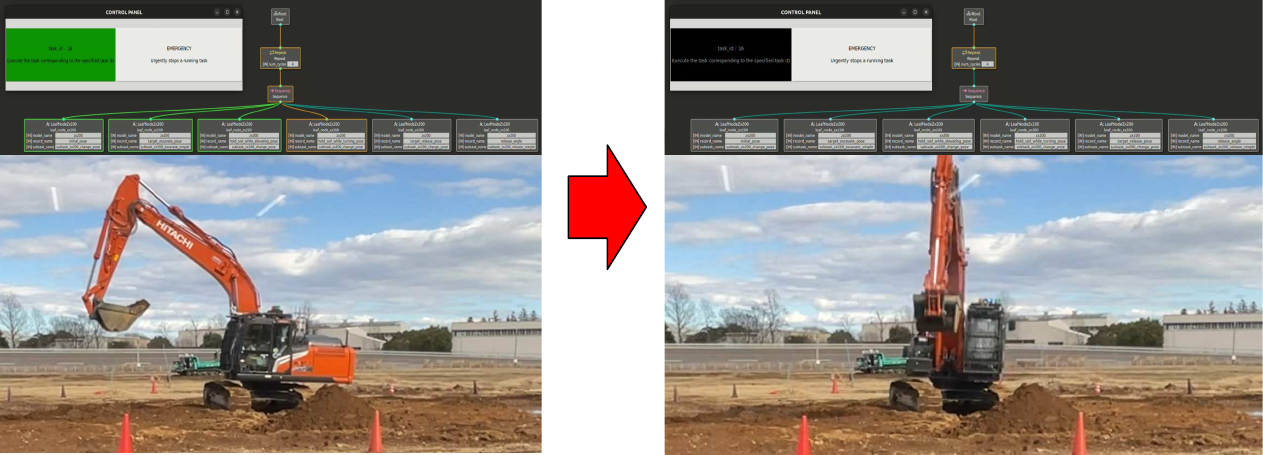}
\caption{Verification results of the task tree and ZX200 during emergency stop operation (Left image: Before emergency stop, Right image: After emergency stop)}
\label{fig:emergency_stop_zx200}
\end{figure}

\begin{figure}[htb]
\centering
\includegraphics[width=75mm]{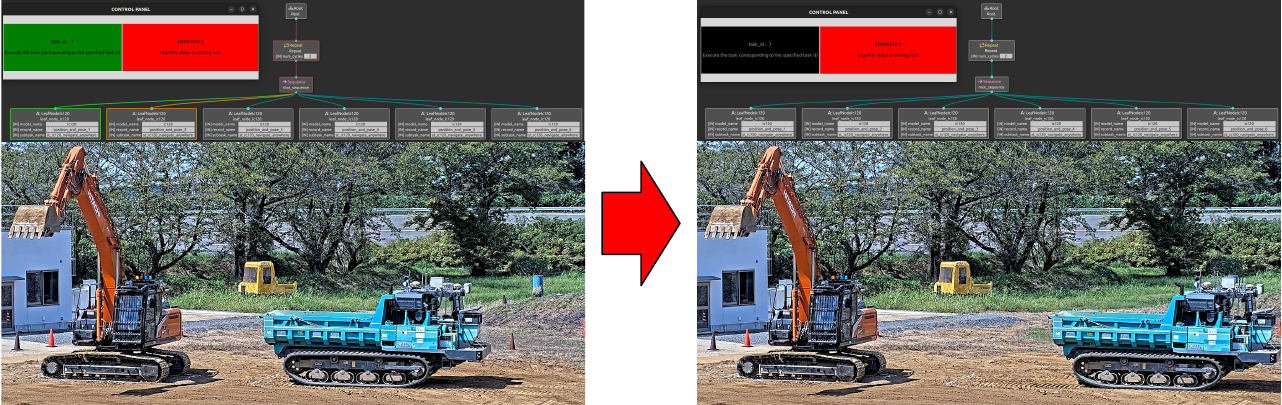}
\caption{Verification results of the task tree and IC120 during emergency stop operation (Left image: Before emergency stop, Right image: After emergency stop)}
\label{fig:emergency_stop_ic120}
\end{figure}

In this way, we carried out the entire series of construction operations and confirmed that everything functioned, as shown in Fig.~\ref{fig:experiment1} and Fig.~\ref{fig:experiment2}. Additionally, we were able to verify that the task cancellation operation works as shown in Fig.~\ref{fig:emergency_stop_zx200} and Fig.~\ref{fig:emergency_stop_ic120}.

\subsection{The experiment of OperaSimVR}
In the test, we connected OperaSimVR to the actual machines. In OperaSimVR, we placed a backhoe and a crawler dump, just like in the real-world field, and imported the terrain data of the experimental field. We also created the surrounding buildings to reproduce the experimental field in cyberspace. For the crawler dump, the global coordinates obtained via GNSS were converted into map coordinates by subtracting the coordinates of a reference point in the experimental field. The parts of the terrain outside the experimental field and the soil mound for excavation were omitted to reduce the load on the PC. In the experiment, we mounted a camera on the actual machine's operator seat to compare the view from the real machine's seat with those in OperaSimVR. To reduce the load on the PC controlling the actual machine, we prepared another PC, and set its ROS domain ID to match that of the actual machine, launched ROS-TCP-Endpoint, and connected it to the system.
\subsection{The result of the experiment on OperaSimVR}
\begin{figure}[htb]
\centering
\includegraphics[width=75mm]{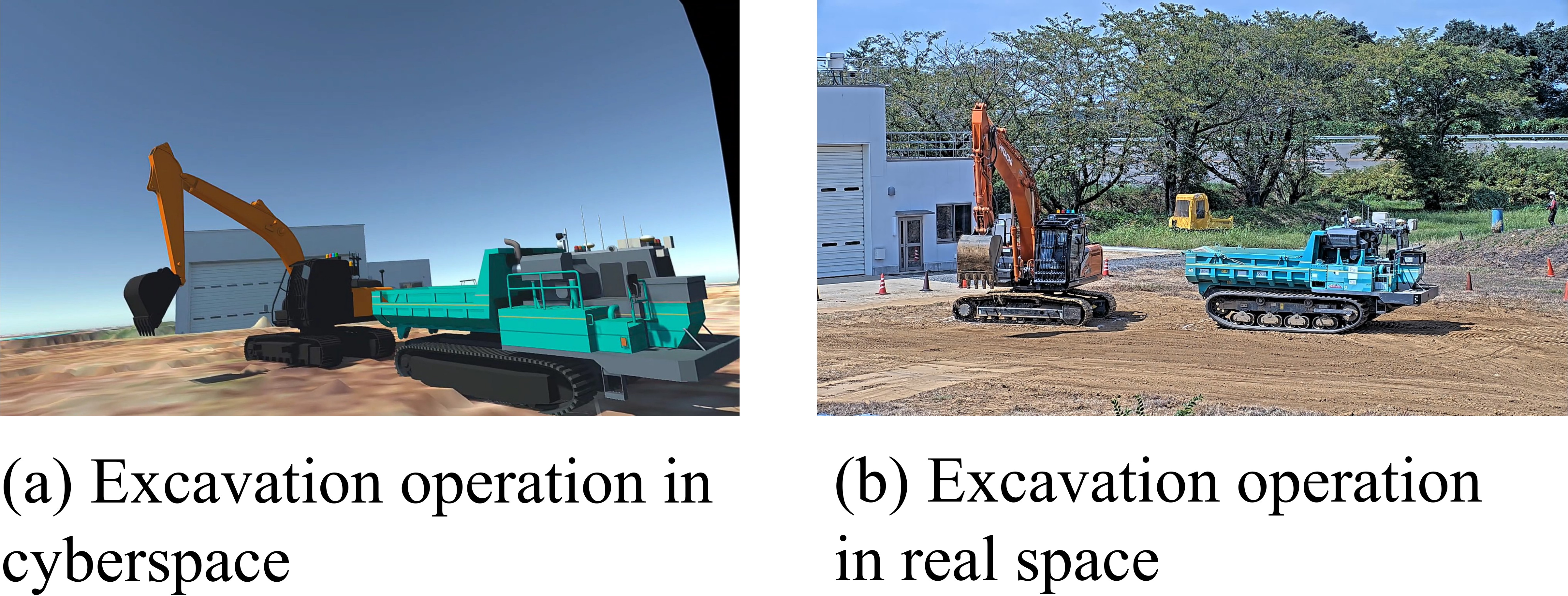}
\caption{Excavation operations in cyberspace and real space.}
\label{fig:vr_2}
\end{figure}

\begin{figure}[htb]
\centering
\includegraphics[width=75mm]{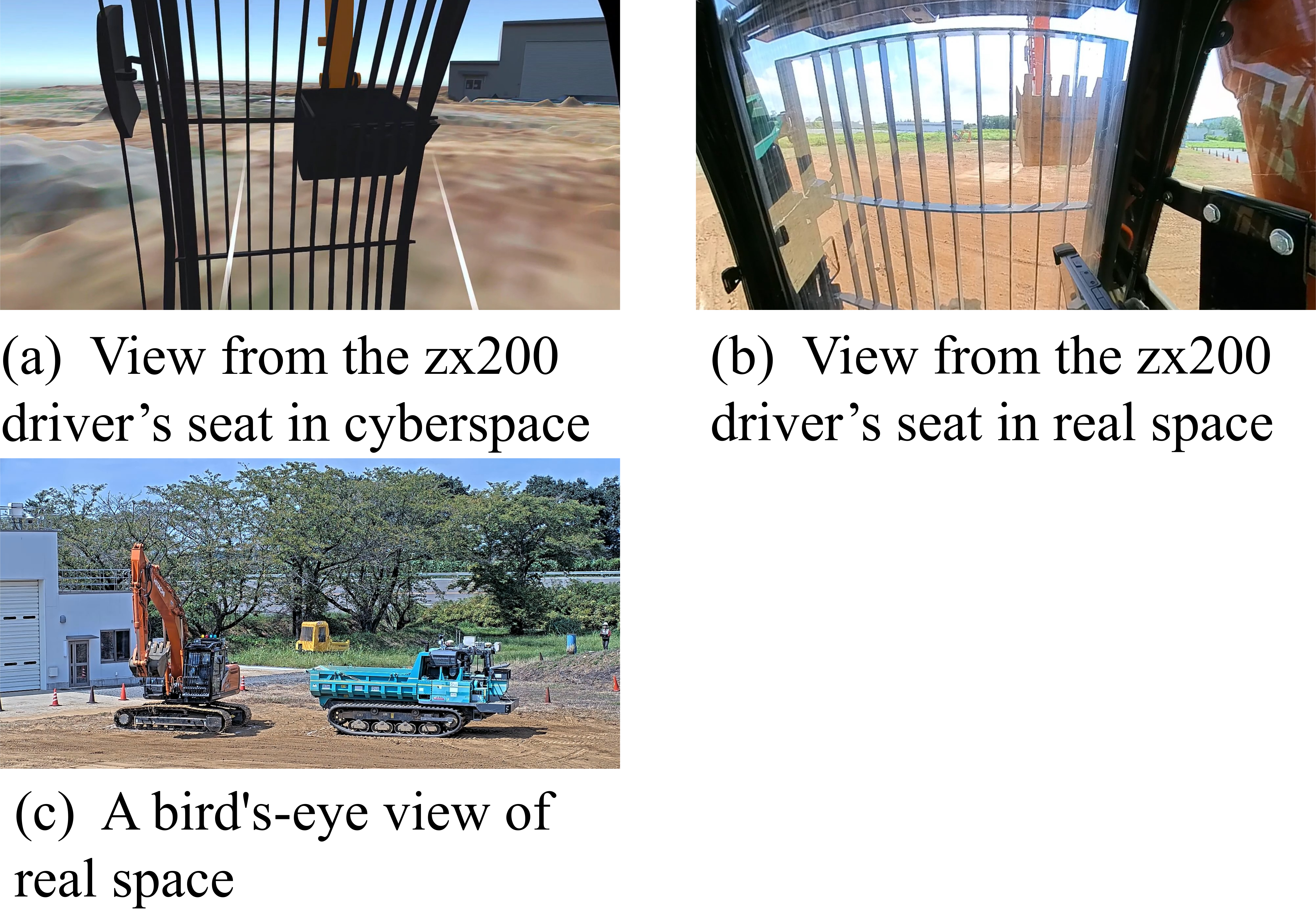}
\caption{Views from the backhoe zx200 driver’s seat in cyberspace and real space during excavation.}
\label{fig:vr_3}
\end{figure}

\begin{figure}[htb]
\centering
\includegraphics[width=75mm]{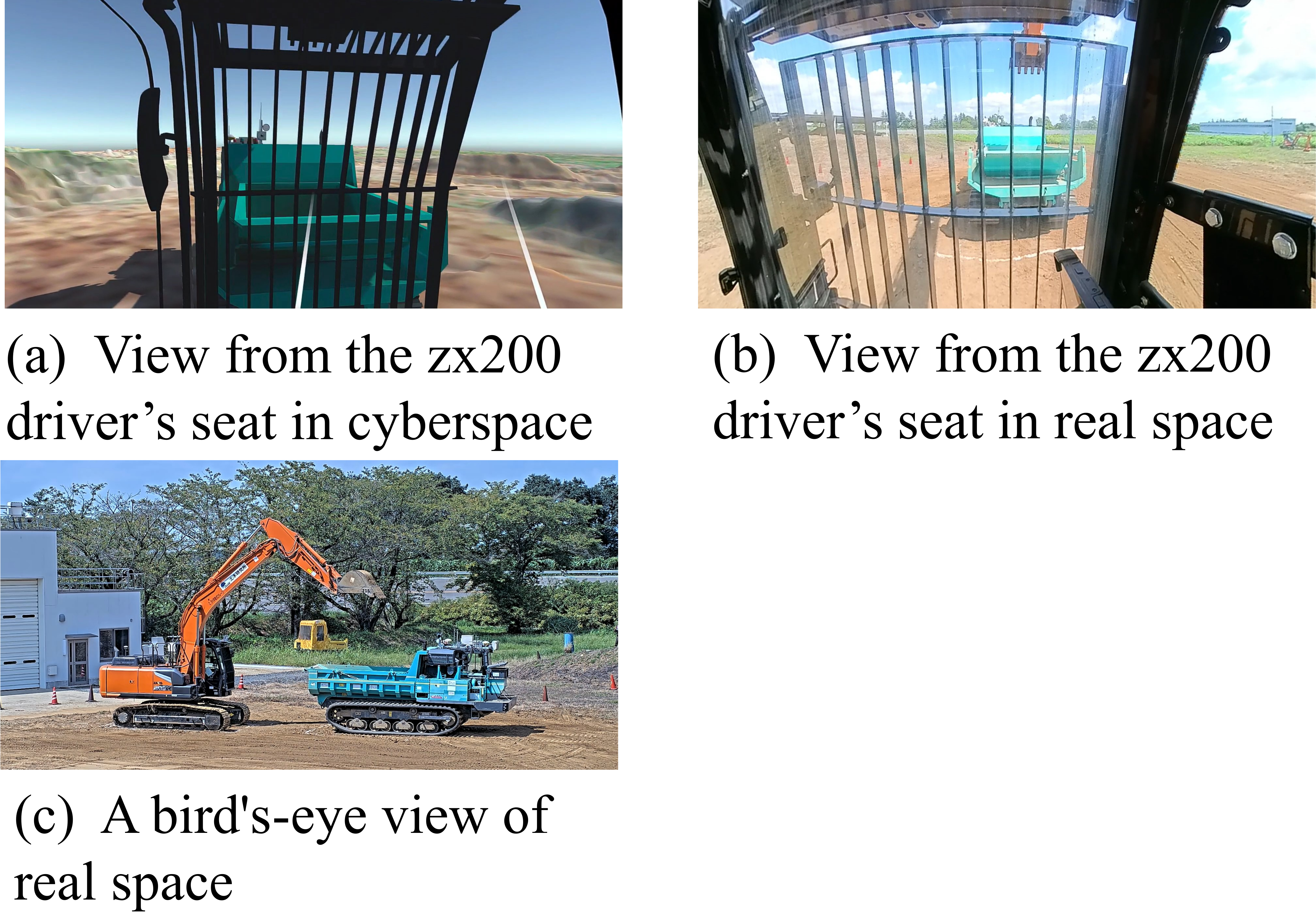}
\caption{Views from the backhoe zx200 driver’s seat in cyberspace and real space during loading.}
\label{fig:vr_4}
\end{figure}

\begin{figure}[htb]
\centering
\includegraphics[width=75mm]{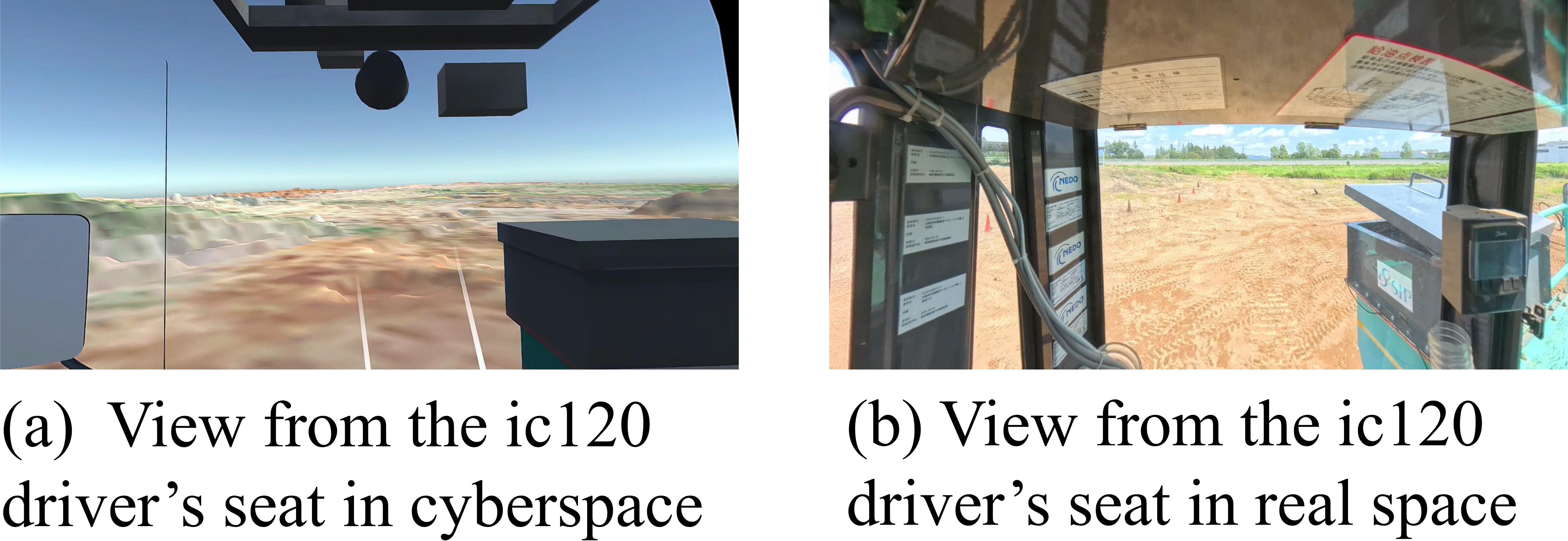}
\caption{Views from the crawler dump ic120 driver’s seat in cyberspace and real space at loading point.}
\label{fig:vr_5}
\end{figure}

Figs.~\ref{fig:vr_2} to \ref{fig:vr_5} show the immersive images of OperaSimVR and images captured by a camera installed in the actual environment during operation. Fig. \ref{fig:vr_2} illustrates the excavating operation in both OperaSimVR and the real world, showing that the positions of the heavy machinery and the posture of the backhoe arm in both cyberspace and the real world are nearly identical. Figs. \ref{fig:vr_3} to \ref{fig:vr_5} present images of the excavation of the soil mound from the operator's seat of the backhoe and crawler dump in both cyberspace and the real world, along with an aerial view from the actual environment. While there were differences in the appearance of the terrain and buildings in cyberspace and the real world due to the terrain data being outdated and the fixed position of the actual camera, they were mostly similar.
When connecting OperaSimVR with the actual machine, there were instances where the system became sluggish, causing the video to freeze while moving through cyberspace. This could be attributed not only to the processing of receiving topics from the actual machine to operate the heavy machinery model but also to the increased load on the PC from loading terrain data and applying textures to the ground and buildings. Additionally, the communication environment was unstable, leading to inconsistent speeds in receiving topics, which may have contributed to the issue. However, there was almost no time delay in the operation of the heavy machinery models between the real world and cyberspace during the connection. When using this technology in the earthwork sites, it is thought that regularly flying drones and updating terrain data from 3D LiDAR installed in the environment could resolve issues related to discrepancies in terrain data.\\
This experiment shows that it was possible to reproduce information obtained from sensors at actual earthwork sites in cyberspace, including the operation of heavy machinery, and provide it to users.

\section{CONCLUSION}
In this study, we introduced ROS2-TMS for Construction constituting of various modules for operating construction machinery coordinately utilizing OPERA. Furthermore, we introduced OperaSimVR, a new feature of ROS2-TMS for Construction, and a new method for implementing tasks in the task management mechanism. Particularly in task implementation, by synchronizing these tasks through a mechanism called the global\_blackboard, we confirmed that more flexible construction operations can be achieved. In addition, we introduced OperaSimVR, one of the main features of ROS2-TMS for Construction that creates a real-time construction site in cyberspace. To validate the proposed mechanisms, we conducted an experiment using the actual crawler dump IC120 and backhoe ZX200, and demonstrated the usability of these mechanisms. Moreover, as illustrated in Fig.~\ref{fig:task_flowchart}, the framework for dynamically updating parameter values based on sensed data has been established, and we plan to integrate additional sensing processes to enable even more flexible autonomous construction. Additionally, other construction machinery in OPERA, including bulldozers, rollers, and dump trucks, as well as machinery compatible with common control signals, is planned to be added. Alongside this, we will begin developing modules for the autonomous control of these machinery via ROS2-TMS for Construction, while also adding the subtasks to support various construction operations. As shown in Table~\ref{tbl:prepared_subtask_nodes}, the current subtasks are still low-level and only involve moving to designated positions and orientation or pose. As construction operations become more complex, it will become increasingly difficult to achieve a sequence of operations using such low-level subtasks alone. What is needed, therefore, is further abstraction of the concept of 'subtasks.' For example, operations like 'excavation and soil dumping' performed by a backhoe could be stored in memory as a task tree, and this task tree could then be used as a 'subtask' when designing new tasks. A key consideration will be the usability of these abstracted subtasks, which must be practical and understandable for civil engineers. Therefore, we will collaborate with civil engineering researchers to design subtasks in the next step. Based on their feedback, we will align our implementation approach and continue to expand functionality.




\section*{Acknowledgements}
This work was supported by Council for Science, Technology and Innovation(CSTI), Cross-ministerial Strategic Innovation Promotion Program (SIP), the 3rd period of SIP “Smart Infrastructure Management System” Grant Number JPJ012187 (Funding agency: Public Works Research Institute). The experiment was conducted in cooperation with Shimizu Corporation.

\end{document}